\pdfoutput=1
\documentclass[10pt,twocolumn,letterpaper]{article}

\usepackage[pagenumbers]{cvpr} 

\usepackage{graphicx, amsmath, amssymb, caption, subcaption, multirow, overpic, textpos}
\usepackage[table]{xcolor}
\usepackage[british, english, american]{babel}
\usepackage[pagebackref=false, breaklinks=true, letterpaper=true, colorlinks,citecolor=citecolor, linkcolor=linkcolor, bookmarks=false]{hyperref}
\definecolor{citecolor}{HTML}{0071BC}
\definecolor{linkcolor}{HTML}{ED1C24}

\def\cvprPaperID{**}
\def\confName{****}
\def\confYear{****}

\usepackage{overpic}
\usepackage{setspace}

\definecolor{baselinecolor}{gray}{.9}
\definecolor{bittersweet}{rgb}{1.0, 0.44, 0.37}
\definecolor{bleudefrance}{rgb}{0.19, 0.55, 0.91}
\usepackage[capitalize]{cleveref}
\crefname{section}{Sec.}{Secs.}
\Crefname{section}{Section}{Sections}
\Crefname{table}{Table}{Tables}
\crefname{table}{Tab.}{Tabs.}

\newlength\savewidth\newcommand\shline{\noalign{\global\savewidth\arrayrulewidth
  \global\arrayrulewidth 1pt}\hline\noalign{\global\arrayrulewidth\savewidth}}
\newcommand{\tablestyle}[2]{\setlength{\tabcolsep}{#1}\renewcommand{\arraystretch}{#2}\centering\footnotesize}
\renewcommand{\paragraph}[1]{\vspace{1.25mm}\noindent\textbf{#1}}

\newcolumntype{x}[1]{>{\centering\arraybackslash}p{#1pt}}
\newcolumntype{y}[1]{>{\raggedright\arraybackslash}p{#1pt}}
\newcolumntype{z}[1]{>{\raggedleft\arraybackslash}p{#1pt}}

\newcommand{\app}{\raise.17ex\hbox{$\scriptstyle\sim$}}

\definecolor{deemph}{gray}{0.6}

\definecolor{baselinecolor}{gray}{.9}

\definecolor{aogreen}{rgb}{0.0, 0.5, 0.0}
\definecolor{bleudefrance}{rgb}{0.19, 0.55, 0.91}
\newcommand{\mbf}[1]{{\mathbf{#1}}}
\def\OURS{FreeNeRF\xspace}

\usepackage{enumitem}
\setlist[itemize]{noitemsep, nolistsep,leftmargin=*}

\begin{document}

\title{FreeNeRF: Improving Few-shot Neural Rendering with Free Frequency Regularization}

\author{Jiawei Yang\\
UC, Los Angeles\\
{\tt\small jiawei118@ucla.edu}
\and
Marco Pavone\\
Nvidia Research, Stanford University\\
{\tt\small pavone@stanford.edu}
\and
Yue Wang\\
Nvidia Research\\
{\tt\small yuewang@nvidia.com}
}

\twocolumn[{%
\maketitle
\centering
\vspace{-1.5em}
\begin{overpic}[width=\linewidth,tics=10]{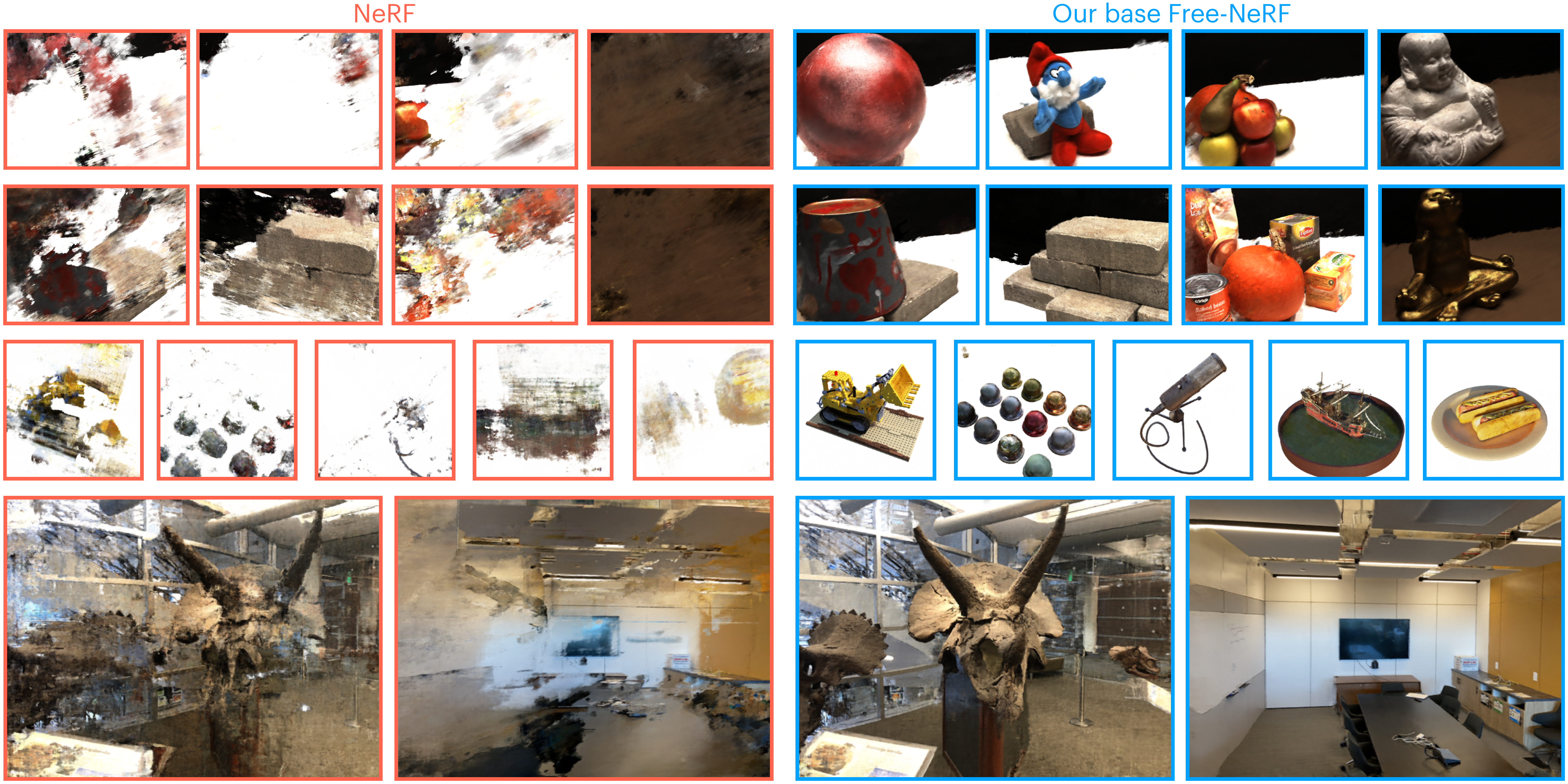}
   \put (0.5,-1.5) {\parbox{\linewidth}{\small 
  Turning {\color{bittersweet} the left} to {\color{bleudefrance} the right} by adding \textbf{\emph{one}} line of code: \texttt{pos\_enc[int(t/T*L)+3:]=0}}}
    \end{overpic}
\captionof{figure}{\textbf{Example novel view synthesis results from sparse inputs.} The only difference between NeRF (left) and FreeNeRF (right) is the use of our frequency regularization, which can be implemented as few as, approximately, \emph{one} line of code (bottom, where $t$ and $T$ denote the current training iteration and regularization duration, respectively; $L$ is the length of the input positional encoding). }
\label{fig:teaser}
\vspace{1em}
}
]
\begin{abstract}
   Novel view synthesis with sparse inputs is a challenging problem for neural radiance fields (NeRF). Recent efforts alleviate this challenge by introducing external supervision, such as pre-trained models and extra depth signals, or by using non-trivial patch-based rendering. In this paper, we present \textbf{Fre}qu\textbf{e}ncy regularized \textbf{NeRF} (FreeNeRF), a surprisingly simple baseline that outperforms previous methods with minimal modifications to plain NeRF. We analyze the key challenges in few-shot neural rendering and find that frequency plays an important role in NeRF’s training. Based on this analysis, we propose two regularization terms: one to regularize the frequency range of NeRF’s inputs, and the other to penalize the near-camera density fields. Both techniques are ``free lunches'' that come at no additional computational cost. We demonstrate that even with just one line of code change, the original NeRF can achieve similar performance to other complicated methods in the few-shot setting. \OURS achieves state-of-the-art performance across diverse datasets, including Blender, DTU, and LLFF. 
   We hope that this simple baseline will motivate a rethinking of the fundamental role of frequency in NeRF's training,  under both the low-data regime and beyond. This project is released at \href{https://jiawei-yang.github.io/FreeNeRF/}{FreeNeRF}.
\end{abstract}

\section{Introduction}
\label{sec:intro}

Neural Radiance Field (NeRF) \cite{mildenhall2020nerf} has gained tremendous attention in 3D computer vision and computer graphics due to its ability to render high-fidelity novel views. However, NeRF is prone to overfitting to training views and struggles with novel view synthesis when only a few inputs are available. We term this view synthesis from sparse inputs problem as a few-shot neural rendering problem.

Existing methods address this challenge using different strategies. Transfer learning methods, \eg, PixelNerf \cite{yu2021pixelnerf} and MVSNeRF \cite{chen2021mvsnerf}, pre-train on large-scale curated multi-view datasets and further incorporate per-scene optimization at test time. Depth-supervised methods \cite{deng2022depth,roessle2022dense} introduce estimated depth as an external supervisory signal, leading to a complex training pipeline. Patch-based regularization methods impose regularization from different sources on rendered patches, \eg, semantic consistency regularization \cite{jain2021putting}, geometry regularization \cite{niemeyer2022regnerf,ehret2022nerf}, and appearance regularization \cite{niemeyer2022regnerf}, all at the cost of computation overhead since an additional, non-trivial number of patches must be rendered during training \cite{jain2021putting,niemeyer2022regnerf,ehret2022nerf}.

In this work, we find that a plain NeRF can work surprisingly well with \emph{none} of the above strategies in the few-shot setting by adding (approximately) as few as \emph{one} line of code (see \cref{fig:teaser}). Concretely, we analyze the common failure modes in training NeRF under a low-data regime. Drawing on this analysis, we propose two regularization terms. One is frequency regularization, which directly regularizes the visible frequency bands of NeRF's inputs to stabilize the learning process and avoid catastrophic overfitting  at the start of training. The other is occlusion regularization, which penalizes the near-camera density fields that cause ``floaters,'' another failure mode in the few-shot neural rendering problem. Combined, we call our method \textbf{Fre}qu\textbf{e}ncy regularized \textbf{NeRF} (\OURS), which is ``free'' in two ways. First, it is dependency-free because it requires neither costly pre-training \cite{yu2021pixelnerf,chen2021mvsnerf,jain2021putting,niemeyer2022regnerf} nor extra supervisory signals \cite{deng2022depth,roessle2022dense}. Second, it is overhead-free as it requires no additional training-time rendering for patch-based regularization \cite{jain2021putting,niemeyer2022regnerf,ehret2022nerf}. 

We consider \OURS a simple baseline (with minimal modifications to a plain NeRF) in the few-shot neural rendering problem, although it already outperforms existing state-of-the-art methods on multiple datasets, including Blender, DTU, and LLFF, at almost no additional computation cost. Our contributions can be summarized as follows:
\begin{itemize}
    \setlength\itemsep{0em}
    \item We reveal the link between the failure of few-shot neural rendering and the frequency of positional encoding, which is further verified by an empirical study and addressed by our proposed method. To our knowledge, our method is the first attempt to address few-shot neural rendering from a frequency perspective. 
    \item We identify another common failure pattern in learning NeRF from sparse inputs and alleviate it with a new occlusion regularizer. This regularizer effectively improves performance and generalizes across datasets.
     \item Combined, we introduce a simple baseline, \OURS, that can be implemented with a few lines of code modification while outperforming previous state-of-the-art methods. Our method is dependency-free and overhead-free, making it a practical and efficient solution to this problem.
\end{itemize}
We hope the observations and discussions in this paper will motivate people to rethink the fundamental role of frequency in NeRF’s positional encoding.

\section{Related Work}

\paragraph{Neural fields.} Neural fields~\cite{neural-fields} use deep neural networks to represent 2D images or 3D scenes as continuous functions. The seminal work, Neural Radiance Fields (NeRF) \cite{mildenhall2020nerf}, has been widely studied and advanced in a variety of applications \cite{barron2021mip,barron2022mip,verbin2022ref,mildenhall2022nerf,park2021nerfies,kobayashi2022decomposing,poole2022dreamfusion}, including novel view synthesis \cite{mildenhall2020nerf,martin2021nerf}, 3D generation \cite{poole2022dreamfusion,jain2022zero}, deformation \cite{park2021nerfies,Pumarola20arxiv_D_NeRF,Rebain20arxiv_derf}, video \cite{Li20arxiv_nsff,Xian20arxiv_stnif,Du20arxiv_nerflow,peng2021neural,li2021neural}. Despite tremendous progress, NeRF still requires hundreds of input images to learn high-quality scene representations; it fails to synthesize novel views with a few input views, \eg, 3, 6, and 9 views, limiting its potential applications in the real world.

\paragraph{Few-shot Neural Rendering.} Many works have attempted to address the challenging few-shot neural rendering problem by leveraging extra information. For instance, external models can be used to acquire normalization-flow regularization \cite{niemeyer2022regnerf}, perceptual regularization \cite{zhang2021ners}, depth supervision \cite{roessle2022dense,deng2022depth,wei2021nerfingmvs}, and cross-view semantic consistency \cite{jain2021putting}. Another thread of works \cite{chibane2021stereo,yu2021pixelnerf,chen2021mvsnerf} attempts to learn transferable models by training on a large, curated dataset instead of using an external model. Recent works argue that geometry is the most important factor in few-shot neural rendering and propose geometry regularization \cite{niemeyer2022regnerf,anonymous2023neural,ehret2022nerf} for better performance. However, these methods require expensive pre-training on tailored multi-view datasets \cite{chibane2021stereo,yu2021pixelnerf,chen2021mvsnerf} or costly training-time patch rendering \cite{jain2021putting,niemeyer2022regnerf,anonymous2023neural,ehret2022nerf}, introducing significant overhead in methodology, engineering implementation, and training budgets. In this work, we show that a plain NeRF can work surprisingly well with minimal modifications (a few lines of code) by incorporating our frequency regularization and occlusion regularization. Unlike most previous methods, our approach maintains the same computational efficiency as the original NeRF.

\paragraph{Frequency in neural representations.} Positional encoding lies at the heart of NeRF's success \cite{mildenhall2020nerf,tancik2020fourier}. Previous studies \cite{tancik2020fourier,sitzmann2020implicit} have shown that neural networks often struggle to learn high-frequency functions from low-dimensional inputs. Encoding inputs with sinusoidal functions of different frequencies can alleviate this issue. Recent works show the benefits of gradually increasing the input frequency in different applications, such as non-rigid scene deformation \cite{park2021nerfies}, bundle adjustment \cite{lin2021barf}, surface reconstruction \cite{wang2022hfneus}, and fitting functions with a wider frequency band \cite{hertz2021sape}. Our work leverages frequency curriculum to tackle the few-shot neural rendering problem. Notably, our approach not only demonstrates the surprising effectiveness of frequency regularization in learning from sparse inputs, but also reveals the failure modes behind this problem and why frequency regularization helps.

\section{Method}

\subsection{Preliminaries}

\paragraph{Neural radiance fields.} A neural radiance field (NeRF) \cite{mildenhall2020nerf} uses a multi-layer perceptron (MLP) to represent a scene as a volumetric density field $\sigma$ and associated RGB values $\mbf{c}$ at each point in the scene. It takes as input a 3D coordinate $\mbf{x}\in\mathbb{R}^3$ and a viewing directional unit vector $\mbf{d}\in\mathbb{S}^2$, and outputs the corresponding density and color. In its most basic form, NeRF learns a continuous function $f_\theta(\mbf{x}, \mbf{d})=(\sigma, \mbf{c})$ where $\theta$ denotes MLP parameters.

\paragraph{Positional encoding.} Directly optimizing NeRF over raw inputs $(\mbf{x}, \mbf{d})$ often leads to difficulties in synthesizing high-frequency details \cite{tancik2020fourier,mildenhall2020nerf}. To address this issue, recent work has used sinusoidal functions with different frequencies to map the inputs into a higher-dimensional space \cite{mildenhall2020nerf}:
\begin{equation}
\label{eq:pos_enc}
	\gamma_L(\mbf{x}) = \left[\sin(\mbf{x}), \cos(\mbf{x}), ..., \sin(2^{L-1}\mbf{x}), \cos(2^{L-1}\mbf{x})\right],
\end{equation}
where $L$ is a hyperparameter that controls the maximum encoded frequency and may differ for coordinates $\mbf{x}$ and directional vectors $\mbf{d}$. A common practice is to concatenate the raw inputs with the frequency-encoded inputs as follows:
\begin{equation}
 \mbf{x}' = [\mbf{x}, \gamma_L(\mbf{x})]
 \label{eq:concat}
\end{equation}
This concatenation is applied to both coordinate inputs and view direction inputs.

\paragraph{Rendering.} To render a pixel in NeRF, a ray $\mbf{r}(t) = \mbf{o} + t \mbf{d}$ is cast from the camera's origin $\mbf{o}$ along the direction $\mbf{d}$ to pass through the pixel, where $t$ is the distance to the origin. Within the near and far bounds $[t_{\mathrm{near}}, t_{\mathrm{far}}]$ of the cast ray, NeRF computes the color of that ray using the quadrature of $K$ sampled points $\mbf{t}_K=\{t_1,\dots, t_K\}$:
\begin{gather}
	\hat{\mbf{c}}(\mbf{r}; \theta, \mbf{t}_K) = \sum_{K} T_k (1-\exp(-\sigma_k(t_{k+1}-t_k))) \mbf{c_k}, \nonumber \\ 
	\textrm{with} \quad T_k = \exp\left(-\sum_{k'<k} \sigma_k'\left(t_{k'+1} - t_{k'}
	\right)\right),
\end{gather}
where $\hat{\mbf{c}}(\mbf{r}; \theta, \mbf{t}_K)$ is the final integrated color. Note that the sampled points $\mbf{t}_K$ are in a near-to-far order, \ie, a point with a smaller index $k$ is closer to the camera's origin.

\begin{figure}[t]\centering
\includegraphics[width=\linewidth]{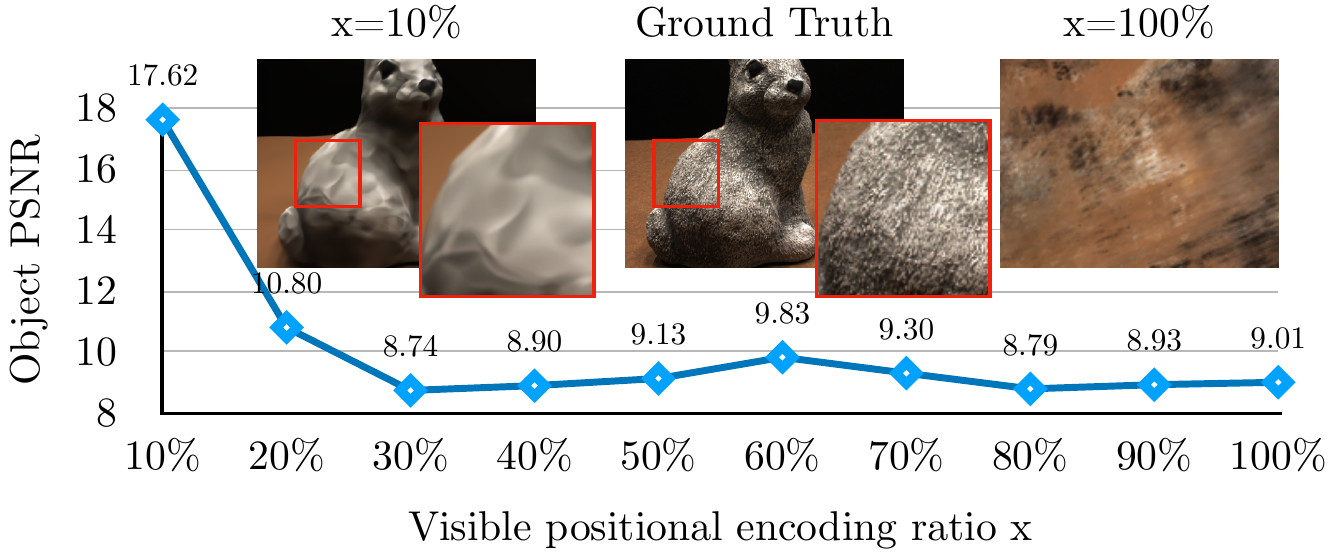}
\caption{\textbf{Masking high-frequency inputs helps few-shot neural rendering.} We investigate how NeRF performs with positional encodings under different masking ratios on the DTU dataset using 3 input views. Despite its over-smoothness, the plain NeRF succeeds in the few-shot setting when only low-frequency inputs are visible.}
\label{fig:masking}
\end{figure}

\subsection{Frequency Regularization}

The most common failure mode of few-shot neural rendering is overfitting. NeRF learns 3D scene representations from a set of 2D images without explicit 3D geometry. 3D geometry is implicitly learned by optimizing appearance in its 2D projected views. However, given only a few input views, NeRF is prone to overfitting to these 2D images with small loss while not explaining 3D geometry in a multi-view consistent way. Synthesizing novel views from such models leads to systematic failure. As shown on the left of \Cref{fig:teaser}, no NeRF model can successfully recover the scene geometry when synthesizing novel views.

The overfitting issue in few-shot neural rendering is presumably exacerbated by high-frequency inputs. \cite{tancik2020fourier} shows that higher-frequency mappings enable faster convergence for high-frequency components. However, the over-fast convergence on high-frequency impedes NeRF from exploring low-frequency information and significantly biases NeRF towards undesired high-frequency artifacts (horns and room examples in \cref{fig:teaser}). In the few-shot scenario, NeRF is even more sensitive to susceptible noise as there are fewer images to learn coherent geometry. Thus, we hypothesize that high-frequency components are a major cause of the failure modes observed in few-shot neural rendering. We provide empirical evidence below.

We investigate how a plain NeRF performs when inputs are encoded by different numbers of frequency bands. To achieve this, we train mipNeRF \cite{barron2021mip} using masked (integrated) positional encoding. Specifically, we set \texttt{pos\_enc[int(L*x\%]):]=0}, where $L$ denotes the length of frequency encoded coordinates after the positional encoding (\cref{eq:pos_enc}), and $x$ is the visible ratio. We briefly demonstrate our observation here and defer the experiment details to \S \ref{sec:setup}. \Cref{fig:masking} shows the results for the DTU dataset under the 3 input-view setting. As anticipated, we observe a significant drop in mipNeRF's performance as higher-frequency inputs are presented to the model. When 10\% of total embedding bits are used, mipNeRF achieves a high PSNR of 17.62, while the plain mipNeRF achieves only 9.01 PSNR on its own (at 100\% visible ratio). The \emph{only} difference between these two models is whether masked positional encodings are used. Although removing a significant portion of high-frequency components avoids catastrophic failure at the start of training, it does not result in competitive scene representations, as the rendered images are usually oversmoothed (as seen in \cref{fig:masking} zoom-in patches). Nonetheless, it is noteworthy that in few-shot scenarios, models using low-frequency inputs may produce significantly better representations than those using high-frequency inputs.

Building on this empirical finding, we propose a frequency regularization method. Given a positional encoding of length $L+3$ (\cref{eq:concat}), we use a linearly increasing frequency mask $\boldsymbol{\alpha}$ to regulate the visible frequency spectrum based on the training time steps, as follows:
\begin{gather}
    \gamma_L'(t,T;\mbf{x}) = \gamma_L(\mbf{x})\odot\boldsymbol{\alpha}(t, T, L), \\
    \resizebox{0.9\linewidth}{!}{$ \displaystyle
        \mathrm{with}~~\boldsymbol{\alpha}_i(t,T,L) = 
        \begin{cases}
            \displaystyle 1 & \text{if}~~ i \leq \frac{t\cdot L}{T} + 3  \\
            \displaystyle \frac{t\cdot L}{T} - \lfloor\frac{t\cdot L}{T}\rfloor & \text{if}~~ \frac{t\cdot L}{T} + 3 < i \leq \frac{t\cdot L}{T} + 6  \\
            0 & \text{if}~~ i > \frac{t\cdot L}{T} + 6
        \end{cases}
    $} \label{eq:mask}
\end{gather}
where $\boldsymbol{\alpha}_i(t,T, L)$ denotes the $i$-th bit value of $\boldsymbol{\alpha}(t,T, L)$; $t$ and $T$ are the current training iteration and the final iteration of frequency regularization, respectively. Concretely, we start with raw inputs without positional encoding and linearly increase the visible frequency by 3-bit each time as training progresses. This schedule can also be simplified as one line of code, as shown in \Cref{fig:teaser}. Our frequency regularization circumvents the unstable and susceptible high-frequency signals at the beginning of training and gradually provides NeRF high-frequency information  to avoid over-smoothness.   

We note that our frequency regularization shares some similarities with the coarse-to-fine frequency schedules used in other works \cite{park2021nerfies,lin2021barf}. Different from theirs, our work focuses on the few-shot neural rendering problem and reveals the catastrophic failure patterns caused by high-frequency inputs and their implication to this problem. 

\subsection{Occlusion Regularization}

\begin{figure}[t]\centering
\includegraphics[width=\linewidth]{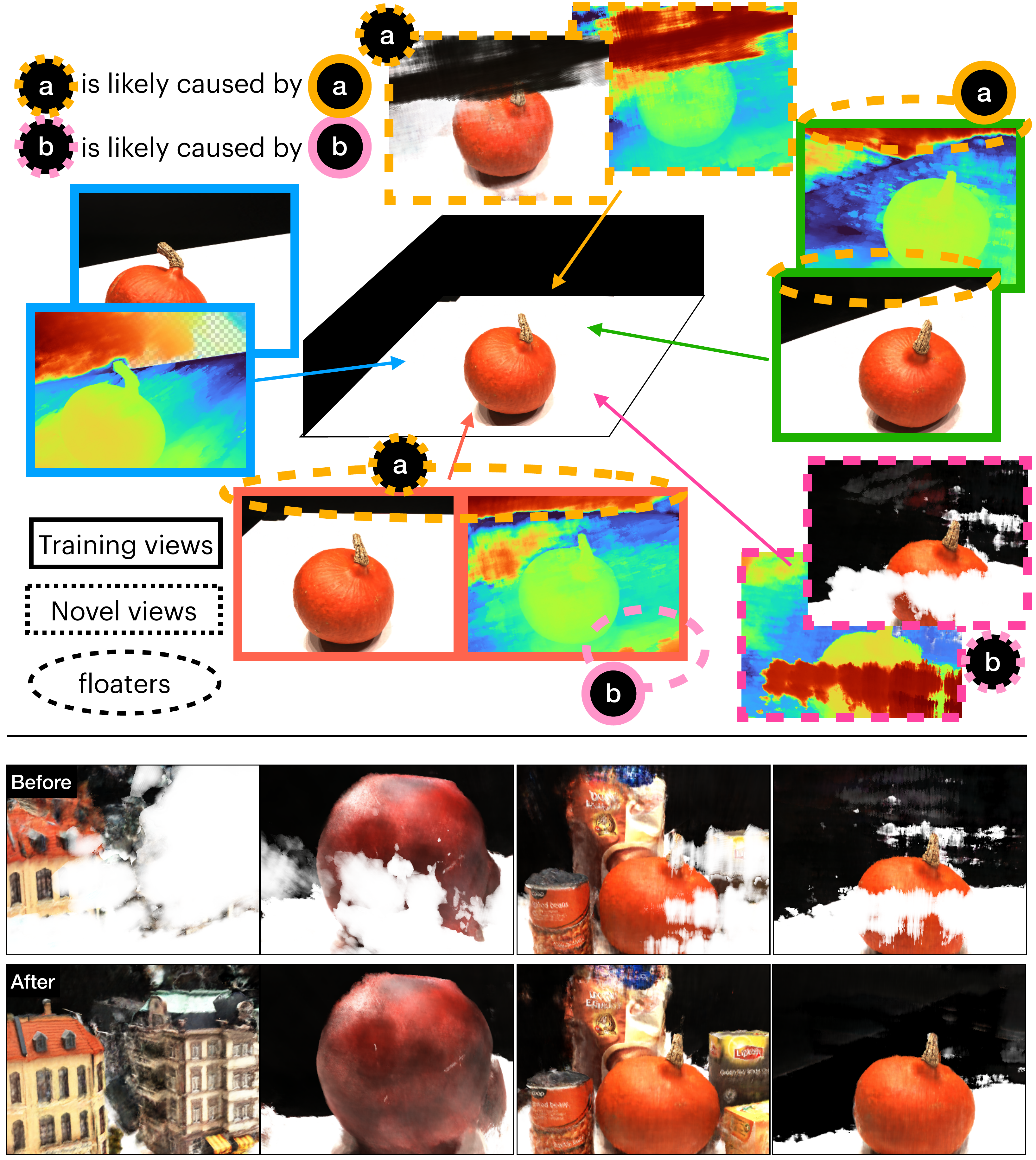}
\caption{\textbf{Illustration of occlusion regularization.} We show 3 training views (solid rectangles) and 2 novel views (dashed rectangles) rendered by a frequency-regularized NeRF. The floaters in the novel views appear to be \emph{near-camera} dense fields in the \emph{training} views (dashed circles) so that we can penalize them directly without the need for the costly novel-view rendering in \cite{jain2021putting,niemeyer2022regnerf}. 
}
\label{fig:occlusion_reg}
\vspace{-1em}
\end{figure}

Frequency regularization does not solve all problems in few-shot neural rendering. Due to the limited number of training views and the ill-posed nature of the problem, certain characteristic artifacts may still exist in novel views. These failure modes often manifest as ``walls'' or ``floaters'' that are located extremely close to the camera, as seen in the bottom of \Cref{fig:occlusion_reg}.  Such artifacts can still be observed even with a sufficient number of training views \cite{barron2022mip}. To address these issues, \cite{barron2022mip} proposed a distortion loss. However, our experiments show that this regularization does not help in the few-shot setting and may even exacerbate the issue. %

We find most of these failure patterns originate from the least overlapped regions in the training views. \Cref{fig:occlusion_reg} shows an example of 3 training views and 2 novel views with ``white walls''. We manually annotate the least overlapped regions in the training views for demonstration ((a) and (b) in \cref{fig:occlusion_reg}). These regions are difficult to estimate in terms of geometry due to the extremely limited information available (one-shot). Consequently, a NeRF model would interpret these unexplored areas as dense volumetric floaters located near the camera. We suspect that the floaters observed in \cite{barron2022mip} also come from these least overlapped regions.

As discussed above, the presence of floaters and walls in novel views is caused by the imperfect training views, and thus can be addressed directly at training time without the need for novel-pose sampling \cite{niemeyer2022regnerf,jain2021putting,yu2021pixelnerf}. To this end, we propose a simple yet effective ``occlusion'' regularization that penalizes the dense fields near the camera. We define:
\begin{equation}
	\mathcal{L}_{occ} = \frac{\boldsymbol{\sigma}_K ^ \intercal \cdot \mbf{m}_K}{K} = \frac{1}{K}\sum_{K} \sigma_k \cdot m_k, %
\end{equation}
where $\mbf{m}_k$ is a binary mask vector that determines whether a point will be penalized, and $\boldsymbol{\sigma}_K$ denotes the density values of the $K$ points sampled along the ray in the order of proximity to the origin (near to far). To reduce solid floaters near the camera, we set the values of $\mbf{m}_k$ up to index $M$, termed as regularization range, to 1 and the rest to 0. The occlusion regularization loss is easy to implement and compute.

\section{Experiments}
\label{sec:exp}

\subsection{Setups} \label{sec:setup}
\paragraph{Datasets \& metrics.} We evaluate our method on three datasets under few-shot settings: the NeRF Blender Synthetic dataset (Blender) \cite{mildenhall2020nerf}, the DTU dataset \cite{jensen2014large}, and the LLFF dataset \cite{mildenhall2019local}. For Blender, we follow DietNeRF \cite{jain2021putting} to train on 8 views and test on 25 test images. For DTU and LLFF, we adhere to RegNeRF's \cite{niemeyer2022regnerf} protocol. On DTU, we use objects' masks to remove the background when computing metrics, as full-image evaluation is biased towards the background, as reported by \cite{yu2021pixelnerf,niemeyer2022regnerf}. We report PSNR, SSIM, and LPIPS scores as quantitative results. We also report the geometric mean of $\mathrm{MSE} = 10^{-\mathrm{PSNR}/10}$, $\sqrt{1 - \mathrm{SSIM}}$, and LPIPS, following  \cite{niemeyer2022regnerf}. More details on the experimental setup can be found in the appendix.

\paragraph{Implementations.} Our \OURS can directly improve NeRF \cite{mildenhall2020nerf} and mipNeRF \cite{barron2021mip}. To demonstrate this, we use DietNeRF's codebase\footnote{\url{https://github.com/ajayjain/DietNeRF}} for NeRF on the Blender dataset and RegNeRF's codebase\footnote{\url{https://github.com/google-research/google-research/tree/master/regnerf}} for mipNeRF on the DTU dataset and the LLFF dataset. We disable the proposed components in those papers and implement our two regularization terms on top of their baselines. We make one modification to mipNeRF \cite{barron2021mip}, which is to concatenate positional encodings with the original Euclidean coordinates (\cref{eq:concat}). This is a default step in NeRF but not in mipNeRF, and it helps unify our experiments’ initial visible frequency range. We follow their training schedules for optimization. Please refer to the Appendix for full training recipes.

\paragraph{Hyper-parameters.} We set the end iteration of frequency regularization as $T=\lfloor 90\%*\mathrm{total\_iters}\rfloor$ for the 3-view setting and $70\%$ for the 6-view setting and $20\%$ for the 9-view setting. We regularize both coordinates $\mbf{x}$ and view directions $\mbf{d}$. For $\mathcal{L}_{occ}$, we use a weight of $0.01$ in all experiments and set the regularization range $M=20$ for LLFF and Blender and  $M=10$ for DTU. For DTU in particular, we find that the ``walls'' are mostly caused by the white desk and black background, so we use this information to penalize more points in a slightly wider range ($M=15$) if their colors are black or white.

\paragraph{Comparing methods.} Unless otherwise specified, we directly use the results reported in DietNeRF \cite{jain2021putting} and RegNeRF \cite{niemeyer2022regnerf} for comparisons, as our method is implemented using their codebases. We also include our reproduced results for reference. %

\begin{table}[ht!]
\centering
\small
\tablestyle{1pt}{1.1}
\begin{tabular}{y{85} | x{40} x{40} x{40}}
Method & PSNR $\uparrow$ & SSIM $\uparrow$ & LPIPS $\downarrow$  \\ \shline
NeRF \cite{mildenhall2020nerf}    & 14.934 & 0.687  & 0.318 \\
NV \cite{lombardi2019neural} &  17.859   & 0.741 & 0.245  \\
Simplified NeRF \cite{jain2021putting}   & 20.092 & 0.822 & 0.179  \\
DietNeRF \cite{jain2021putting} & \cellcolor{yellow!25} 23.147     & \cellcolor{yellow!25} 0.866   & \cellcolor{yellow!25} 0.109  \\
DietNeRF + $\mathcal{L}_\text{MSE}$ ft $50k$  & \cellcolor{orange!25} 23.591 & \cellcolor{orange!25} 0.874 & \cellcolor{red!25} 0.097 \\ \hline
NeRF (repro.) & \cellcolor{gray!25} 13.931 & \cellcolor{gray!25} 0.689 &  \cellcolor{gray!25} 0.320 \\
DietNeRF (repro.) &  22.503          & 0.823   & 0.124  \\
\textbf{Our \OURS} & \cellcolor{red!25} 24.259 & \cellcolor{red!25} 0.883 &  \cellcolor{orange!25} 0.098 \\
\end{tabular}
\vspace{-2mm}
\caption{\textbf{Quantitative comparison on Blender.} ``$\mathcal{L}_\text{MSE}$ ft $50k$'': fine-tune for another $50k$ iterations with $\mathcal{L}_\text{MSE}$. The top row section includes results from \cite{jain2021putting}, while the bottom row section shows our reproduced results (repro.). Gray: our baseline. Red, orange, and yellow: the best, second-best, and third-best.}
\label{tab:blender}
\vspace{-2mm}
\end{table}

\begin{figure}[t!]
\centering
\includegraphics[width=\linewidth]{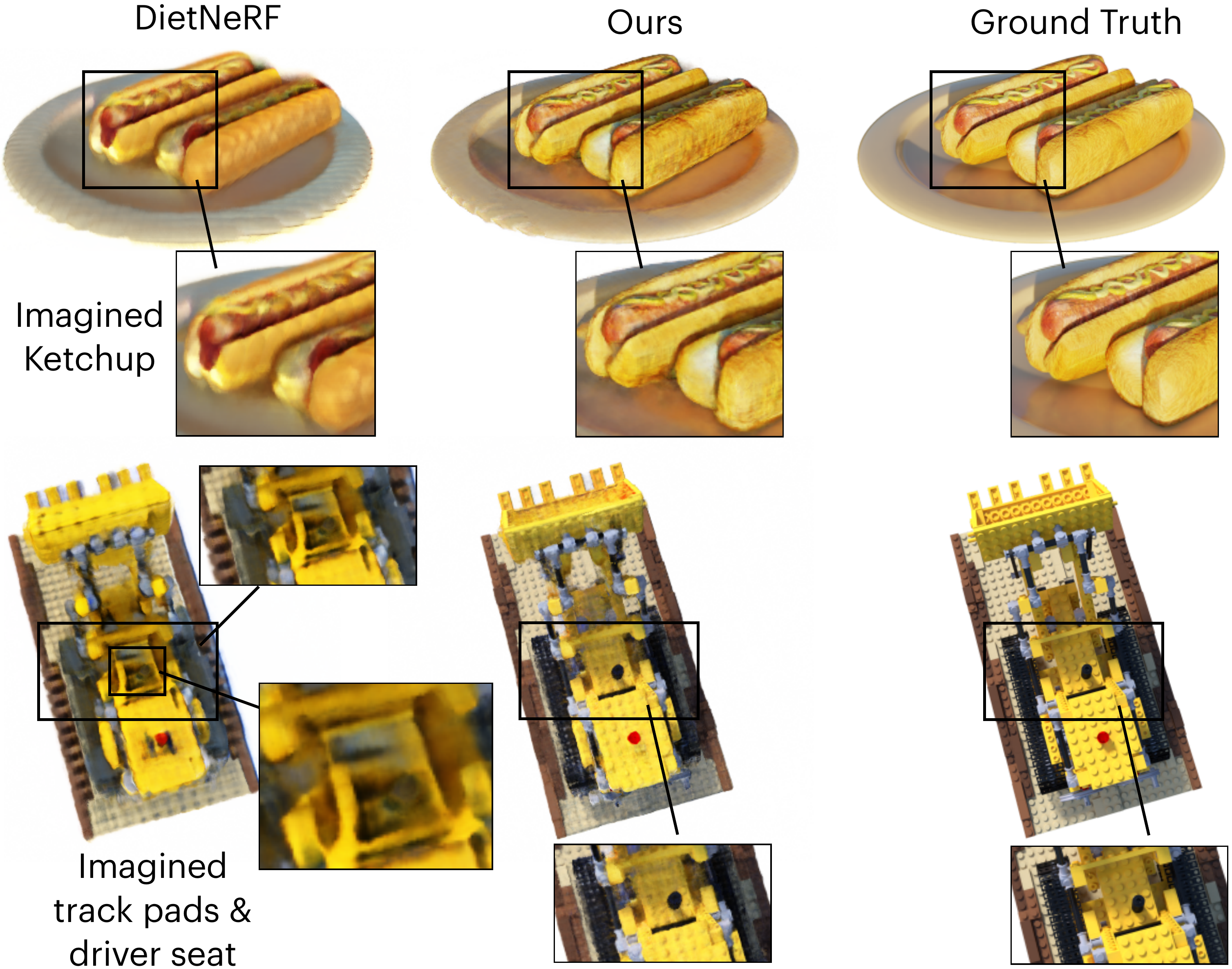}
\caption{\textbf{Novel view synthesis examples on Blender.} Our results are qualitatively better than DietNeRF's. DietNeRF renders ``imaginary'' components that do not exist in the original images.}
\label{fig:dietnerf_imagines}
\end{figure}

\begin{table*}
\centering
\small
\tablestyle{2pt}{1.1}
\begin{tabular}{l | c | x{24}x{24}x{24} | x{24}x{24}x{24} | x{24}x{24}x{24} | x{24}x{24}x{24} }
& \multirow{2}{*}{Setting} & \multicolumn{3}{c|}{Object PSNR $\uparrow$} & \multicolumn{3}{c|}{Object SSIM $\uparrow$} & \multicolumn{3}{c|}{Full-image PSNR $\uparrow$} & \multicolumn{3}{c}{Full-image SSIM $\uparrow$}  \\
  &  & 3-view & 6-view & 9-view  & 3-view & 6-view & 9-view  & 3-view & 6-view & 9-view  & 3-view & 6-view & 9-view \\ \shline
SRF~\cite{chibane2021stereo} & \multirow{3}{*}{Trained on DTU} & 15.32 & 17.54 & 18.35 & 0.671 & 0.730 & 0.752 & 15.84 & 17.77 & 18.56 & 0.532 & 0.616 & 0.652 \\
PixelNeRF~\cite{yu2021pixelnerf} &  & 16.82 & 19.11 & 20.40 & 0.695 & 0.745 & 0.768 & \cellcolor{red!25} 18.74 & \cellcolor{yellow!25} 21.02 & 22.23 & \cellcolor{yellow!25}0.618 & 0.684 & 0.714 \\
MVSNeRF~\cite{chen2021mvsnerf} &  &  18.63 &  20.70 & 22.40 &  \cellcolor{orange!25}0.769 &  0.823 &  0.853 &  16.33 &  18.26 &  20.32 &  0.602 &  0.695 & 0.735 \\
\hline 
SRF ft~\cite{chibane2021stereo} & \multirow{3}{*}{\shortstack{Trained on DTU\\and\\Optimized per Scene}} & 15.68 & 18.87 & 20.75 & 0.698 & 0.757 & 0.785 & 16.06 & 18.69 & 19.97 & 0.550 & 0.657 & 0.678  \\
PixelNeRF ft~\cite{yu2021pixelnerf} &  &  \cellcolor{orange!25}18.95 & 20.56 & 21.83 & 0.710 & 0.753 & 0.781 & \cellcolor{yellow!25} 17.38 & \cellcolor{orange!25} 21.52 & 21.67 & 0.548 & 0.670 & 9.680 \\
MVSNeRF ft~\cite{chen2021mvsnerf} &  & 18.54 & 20.49 & 22.22 &  \cellcolor{orange!25}0.769 &  0.822 &  0.853 &  16.26 & 18.22 & 20.32 & 0.601 & 0.694 & 0.736 \\
\hline 
mip-NeRF~\cite{barron2021mip} & \multirow{3}{*}{Optimized per Scene} & 8.68 & 16.54 &  23.58 & 0.571 & 0.741 &   0.879 & 7.64 & 14.33 & 20.71 & 0.227 & 0.568 & 0.799\\
DietNeRF~\cite{jain2021putting} &  & 11.85 &  20.63 &  23.83 & 0.633 & 0.778 & 0.823 & 10.01 & 18.70 & 22.16 & 0.354 & 0.668 & 0.740 \\
RegNeRF~\cite{niemeyer2022regnerf} &  & \cellcolor{yellow!25} 18.89 &  \cellcolor{orange!25}22.20 &  \cellcolor{orange!25}24.93 &  0.745 &  \cellcolor{yellow!25}0.841 &  \cellcolor{yellow!25}0.884 &  15.33 & 19.10 & \cellcolor{yellow!25} 22.30 & \cellcolor{orange!25} 0.621 & \cellcolor{orange!25} 0.757 & \cellcolor{yellow!25} 0.823 \\ \hline

mip-NeRF concat. (repro.) & \multirow{3}{*}{Optimized per Scene} & \cellcolor{gray!25} 9.10 & \cellcolor{gray!25} 16.84 & \cellcolor{gray!25} 23.56 & \cellcolor{gray!25} 0.578&  \cellcolor{gray!25} \cellcolor{yellow!25} 0.754& \cellcolor{gray!25} 0.877 & \cellcolor{gray!25} 7.94 & \cellcolor{gray!25} 14.15 & \cellcolor{gray!25} 20.97 & \cellcolor{gray!25} 0.235 & \cellcolor{gray!25} 0.560 & \cellcolor{gray!25} 0.794\\
$^\dagger$RegNeRF concat. (repro.) &   & 18.50 &\cellcolor{yellow!25} 22.18 & \cellcolor{yellow!25} 24.88 & 0.744  & \cellcolor{orange!25} 0.844 & \cellcolor{red!25} 0.890 & 15.00 & 19.12 & \cellcolor{orange!25} 22.41  & 0.606 & 0.754 & \cellcolor{orange!25} 0.826  \\
\textbf{Our \OURS} &  & \cellcolor{red!25} 19.92 & \cellcolor{red!25} 23.25 & \cellcolor{red!25} 25.38 & \cellcolor{red!25} 0.787 & \cellcolor{red!25} 0.844 & \cellcolor{orange!25} 0.888 & \cellcolor{orange!25} 18.02 &  \cellcolor{red!25} 22.39 & \cellcolor{red!25} 24.2 & \cellcolor{red!25} 0.680 &\cellcolor{red!25} 0.779 & \cellcolor{red!25} 0.833 \\
\end{tabular}
\vspace{-1em}
\caption{
    \textbf{Quantitative comparison on DTU.} We present the PSNR and SSIM scores of foreground objects and full images. Our \OURS synthesizes better foreground objects and full images than most of the others. Our direct baseline is mipNeRF \cite{barron2021mip} (marked in {gray}). Results in the bottom row section are our reproductions, and others come from \cite{niemeyer2022regnerf}. ``concat.'': inputs concatenation (\cref{eq:concat}). $^\dagger$ReNeRF: w/o. appearance regularization. The best, second-best, and third-best entries are marked in red, orange, and yellow, respectively.}
\vspace{-1em}
\label{tab:dtu}
\end{table*}
\begin{figure*}[t!]\centering
\includegraphics[width=\linewidth]{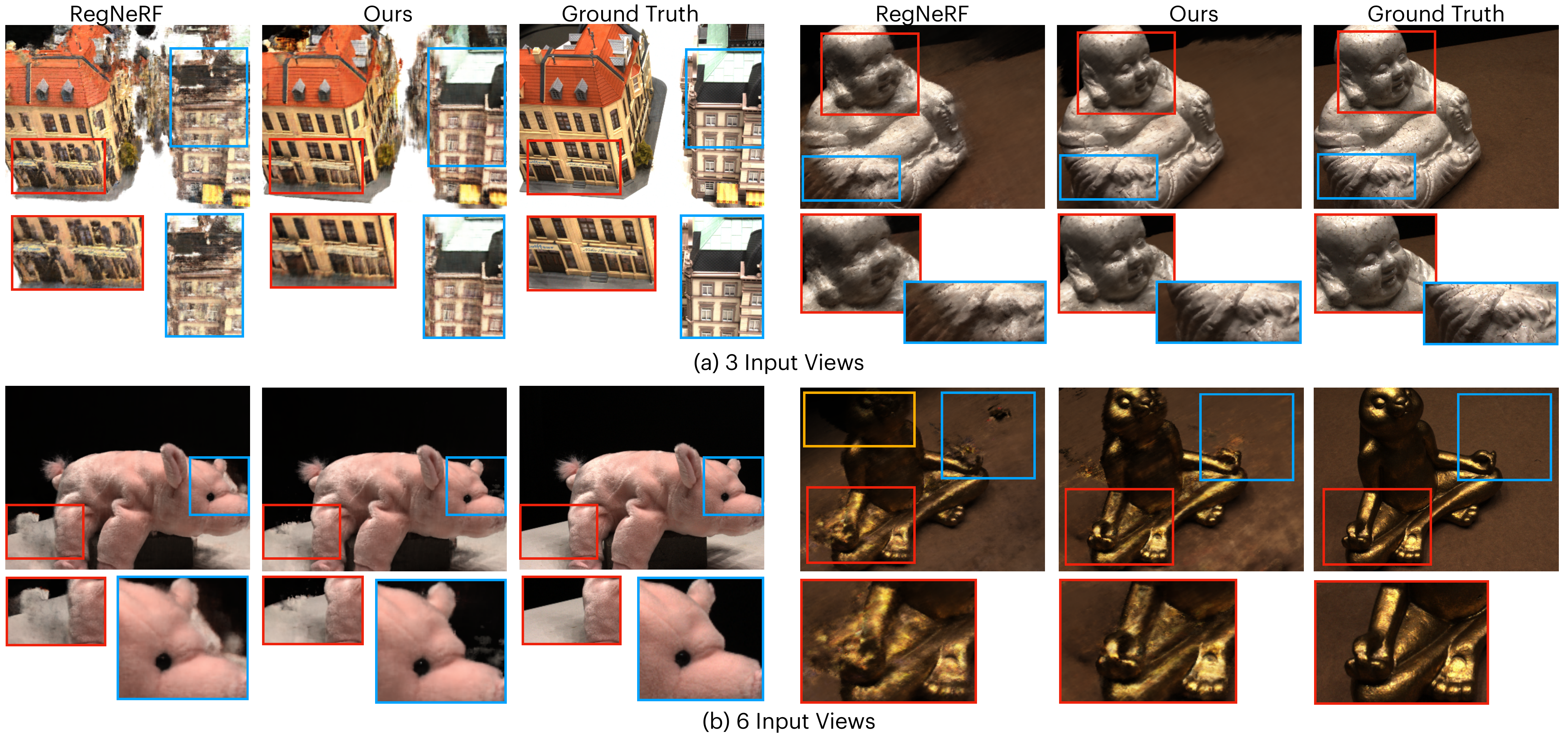}
\vspace{-2em}
\caption{\textbf{Qualitative comparison on DTU.} We show novel views rendered by RegNeRF and ours in 3 and 6 input-view settings. For the Buddha example, the piece-wise geometry regularization used by RegNeRF \cite{niemeyer2022regnerf} hurts the fine-grained geometry, erasing the details of eyes, fingers and wrinkles. RegNeRF's results are rendered by our reproduced $^\dagger$RegNeRF concat. (\emph{c.f.}~\cref{tab:dtu}). 
}
\vspace{-1em}
\label{fig:dtu}
\end{figure*}

\subsection{Comparison}

We compare with state-of-the-art methods in terms of novel view synthesis quality and computation overhead. We show that \OURS outperforms others in synthesis quality while maintaining a much lower cost.

\begin{table*}
\centering
\small
\tablestyle{2pt}{1.1}
\begin{tabular}{l | c | x{24}x{24}x{24} | x{24}x{24}x{24} | x{24}x{24}x{24} | x{24}x{24}x{24} }
& \multirow{2}{*}{Setting} & \multicolumn{3}{c|}{PSNR $\uparrow$} & \multicolumn{3}{c|}{SSIM $\uparrow$} & \multicolumn{3}{c|}{LPIPS $\downarrow$} & \multicolumn{3}{c}{Average $\downarrow$}  \\
  &  & 3-view & 6-view & 9-view  & 3-view & 6-view & 9-view  & 3-view & 6-view & 9-view  & 3-view & 6-view & 9-view \\ \shline
SRF~\cite{chibane2021stereo} & \multirow{3}{*}{Trained on DTU} & 12.34 & 13.10 & 13.00 & 0.250 & 0.293 & 0.297 & 0.591 & 0.594 & 0.605 & 0.313 & 0.293 & 0.296 \\
PixelNeRF~\cite{yu2021pixelnerf} &  & 7.93 & 8.74 & 8.61 & 0.272 & 0.280 & 0.274 & 0.682 & 0.676 & 0.665 & 0.461 & 0.433 & 0.432 \\
MVSNeRF~\cite{chen2021mvsnerf} &  & 17.25 & 19.79 & 20.47 & 0.557 & 0.656 & 0.689 & 0.356 & 0.269 & 0.242 & 0.171 & 0.125 & 0.111 \\
\hline 
SRF ft~\cite{chibane2021stereo} & \multirow{3}{*}{\shortstack{Trained on DTU\\and\\Optimized per Scene}} & 17.07 & 16.75 & 17.39 & 0.436 & 0.438 & 0.465 & 0.529 & 0.521 & 0.503 & 0.203 & 0.207 & 0.193 \\
PixelNeRF ft~\cite{yu2021pixelnerf} &  & 16.17 & 17.03 & 18.92 & 0.438 & 0.473 & 0.535 & 0.512 & 0.477 & 0.430 & 0.217 & 0.196 & 0.163 \\
MVSNeRF ft~\cite{chen2021mvsnerf} &  & 17.88 & 19.99 & 20.47 & \cellcolor{yellow!25}0.584 & 0.660 & 0.695 & \cellcolor{orange!25}0.327 & 0.264 & 0.244 & 0.157 & 0.122 & 0.111 \\
\hline 
mip-NeRF~\cite{barron2021mip} & \multirow{3}{*}{Optimized per Scene} & 14.62 & 20.87 & 24.26 & 0.351 &0.692 & 0.805 & 0.495 & 0.255 & 0.172 & 0.246 & 0.114 & 0.073 \\
DietNeRF~\cite{jain2021putting} &  & 14.94 & 21.75 & 24.28 & 0.370 & 0.717 & 0.801 & 0.496 & 0.248 & 0.183 & 0.240 & 0.105 & 0.073 \\
RegNeRF~\cite{niemeyer2022regnerf} &  & \cellcolor{orange!25}19.08 & \cellcolor{yellow!25} 23.10 & 24.86 & \cellcolor{orange!25}0.587 & \cellcolor{yellow!25}0.760 & \cellcolor{yellow!25}0.820 & \cellcolor{yellow!25}0.336 & \cellcolor{yellow!25}0.206 & \cellcolor{yellow!25}0.161 & \cellcolor{orange!25}0.149 & \cellcolor{orange!25}0.086 & 0.067 \\ \hline

mip-NeRF concat. (repro.) & \multirow{3}{*}{Optimized per Scene} & \cellcolor{gray!25} 16.11 & \cellcolor{gray!25} 22.91  & \cellcolor{orange!25} 24.88  & \cellcolor{gray!25} 0.401 &  \cellcolor{gray!25} 0.756 & \cellcolor{orange!25} 0.826 & \cellcolor{gray!25} 0.460  & \cellcolor{gray!25} 0.213  &  \cellcolor{orange!25} 0.160 & \cellcolor{gray!25}0.215 & \cellcolor{gray!25} 0.090& \cellcolor{orange!25} 0.066\\
$^\dagger$RegNeRF concat. (repro.) &   & \cellcolor{yellow!25}18.84 & \cellcolor{orange!25} 23.22 & \cellcolor{yellow!25} 24.88  & 0.573 & \cellcolor{orange!25} 0.770  & \cellcolor{orange!25} 0.826  & 0.345 & \cellcolor{orange!25} 0.203 & \cellcolor{red!25}0.159 & \cellcolor{yellow!25} 0.150 & \cellcolor{yellow!25} 0.085 & \cellcolor{yellow!25}0.065 \\
\textbf{Our \OURS} &  & \cellcolor{red!25} 19.63 & \cellcolor{red!25} 23.73 & \cellcolor{red!25} 25.13 & \cellcolor{red!25} 0.612 & \cellcolor{red!25} 0.779 & \cellcolor{red!25}  0.827 & \cellcolor{red!25} 0.308 &  \cellcolor{red!25} 0.195 & \cellcolor{orange!25} 0.160 & \cellcolor{red!25} 0.134 &\cellcolor{red!25} 0.075& \cellcolor{red!25}0.064 \\
\end{tabular}
\vspace{-1em}
    \caption{
    \textbf{Quantitative comparison on LLFF.} Our \OURS achieves the best results in most metrics under different input-view settings. 
    Our direct baseline is mipNeRF \cite{barron2021mip} (marked in {gray}). Results in the bottom row section are our reproductions, and others come from \cite{niemeyer2022regnerf}. ``concat.'': inputs concatenation (\cref{eq:concat}). $^\dagger$ReNeRF: w/o. appearance regularization. The best, second-best, and third-best entries are marked in red, orange, and yellow, respectively.
    }
    \label{tab:llff}
    \vspace{-1.5em}
\end{table*}

\begin{figure*}[ht!]\centering
\includegraphics[width=\linewidth]{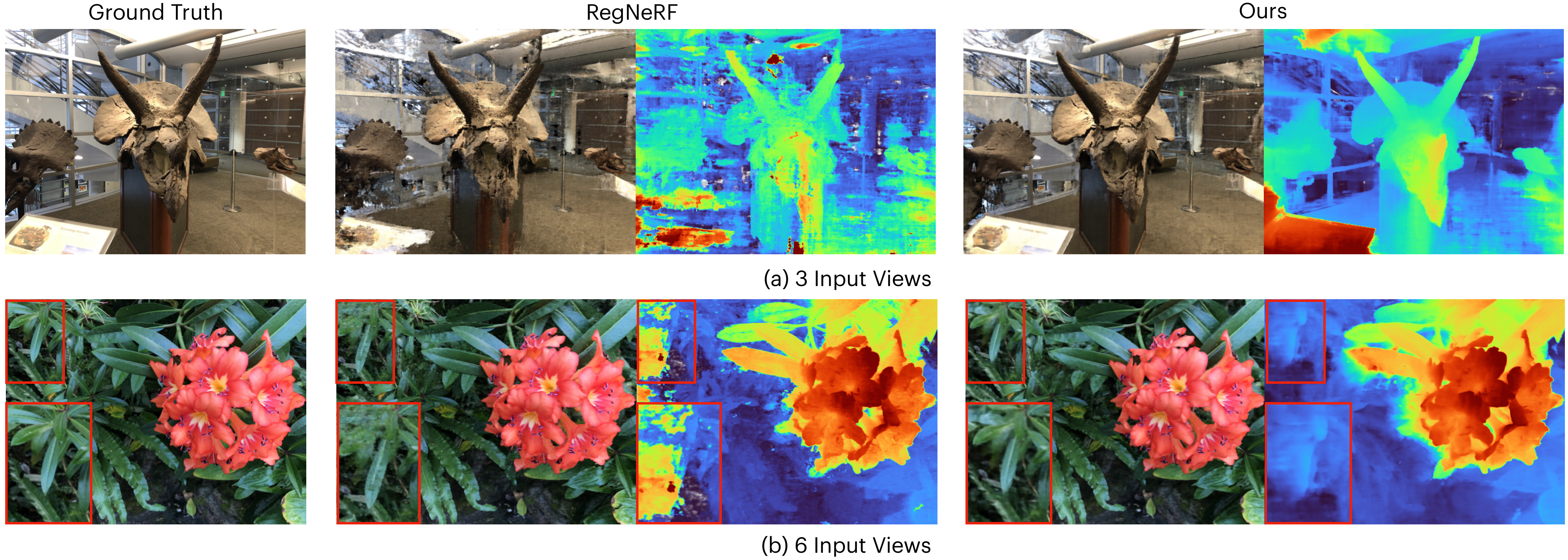}
\vspace{-2em}
\caption{\textbf{Qualitative comparison on LLFFF.} 
RegNeRF \cite{niemeyer2022regnerf} fails to estimate the accurate depth though it renders visually satisfactory RGB images (a). It also suffers from near-camera floaters (b). In contrast, our method reconstructs less noisy occupancy fields with fewer floaters. RegNeRF's results are rendered by our reproduced $^\dagger$RegNeRF concat. (\emph{c.f.}~\cref{tab:llff}). 
}
\vspace{-1em}
\label{fig:llff}
\end{figure*}

\paragraph{Blender dataset.} \Cref{tab:blender} shows the image synthesis metrics on the Blender dataset \cite{mildenhall2020nerf}. Our approach outperforms all other methods in the PSNR and SSIM scores, with a comparable LPIPS score to the best one. The improved DietNeRF with fine-tuning still underperforms ours. Note that our direct baseline is ``NeRF (repro.)'' as we do not use any techniques from DietNeRF \cite{jain2021putting}.
\Cref{fig:dietnerf_imagines} shows two examples for qualitative comparison (see \cref{fig:teaser} for plain NeRF's results). Interestingly, we observe that DietNeRF implicitly distills semantic information from a pre-trained CLIP model \cite{radford2021learning} into NeRF, which leads to unrealistic and ``imaginary'' patches that do not exist in the original scenes, such as ``ketchup'' in the hotdog and rubber-like track-pads in the bulldozer. This behavior is highly correlated to feature distillation \cite{kobayashi2022decomposing} and recent developments in 3D object generation that combine NeRF with large pre-trained vision-language models \cite{jain2022zero,poole2022dreamfusion}. Although this potentially could be an interesting application, such behavior is undesired in our task and will hamper outputs' fidelity. In contrast, our method does not require semantics regularization while achieving better performance.

\paragraph{DTU dataset.} \Cref{tab:dtu} shows the quantitative results on the DTU dataset. 
Transfer learning-based methods that require expensive pre-training (SRF \cite{chibane2021stereo}, PixelNeRF\cite{yu2021pixelnerf}, and MVSNeRF \cite{chen2021mvsnerf}) underperform ours in almost all settings, except the full-image PSNR score under 3-view setting. This may be due to the bias introduced by the white table and black background present in many scenes in the DTU dataset, which can be learned as a prior through pre-training. Compared to per-scene optimization methods (mipNeRF \cite{barron2021mip}, DietNeRF \cite{jain2021putting}, and RegNeRF \cite{niemeyer2022regnerf}), our approach achieves the best results. \Cref{fig:dtu} shows example novel views rendered by RegNeRF and ours. In the Buddha scene, for instance, piece-wise smoothness imposed by RegNeRF's geometry regularization \cite{niemeyer2022regnerf} leads to the loss of fine-grained details, such as eyes, fingers, and wrinkles. In contrast, our frequency regularization, which can be seen as an implicit geometry regularization, forces smooth geometry at the beginning (due to the limited frequency spectrum) and gradually relaxes the constraint to facilitate the details. In the more challenging scenes (\eg, buildings and the bronze statue in \cref{fig:dtu}), \OURS produces higher-quality results.

\paragraph{LLFF dataset.} \Cref{tab:llff} and \Cref{fig:llff} show quantitative and qualitative results, respectively, on the LLFF dataset. We reproduce mipNeRF \cite{barron2021mip} and obtain better results. Our \OURS is generally the best. Transfer learning-based methods \cite{chibane2021stereo, chen2021mvsnerf, yu2021pixelnerf} perform much worse than ours on the LLFF dataset due to the non-trivial domain gap between DTU and LLFF. Compared to RegNeRF \cite{niemeyer2022regnerf}, our approach predicts more precise geometry and exhibits fewer artifacts. For instance, RegNeRF's rendered ``horns'' example (\cref{fig:llff}-a) is perceptually acceptable but has poor depth map quality, indicating its incorrect geometry estimation. \OURS, in contrast, renders a less noisy and smoother occupancy field. Also, our approach suffers less from ``floaters'' than ReNeRF (\cref{fig:llff}-b), further demonstrating the efficacy of our occlusion regularization.

\paragraph{Training overhead.} In \Cref{tab:overhead}, we include the training time of different methods under the same setting. Our method only introduces negligible training overhead ($1.02-1.04\times$) compared to the other approaches ($1.62-2.8\times$). Both DietNeRF \cite{jain2021putting} and RegNeRF \cite{niemeyer2022regnerf} render unobserved patches from novel poses for regularization, which significantly sets back the training efficiency. DietNeRF requires additional forward evaluation of a large model (CLIP ViT B/32, $224^2$, \cite{radford2021learning}), and RegNeRF also experiences increased computation due to the use of a normalizing flow model (this part is not open-sourced and therefore not available for our experiments). In contrast, \OURS does not require such additional steps, making it a lightweight and efficient solution for addressing few-shot neural rendering problems.

\begin{table}[ht!]
\tablestyle{2pt}{1.1}
\begin{tabular}{x{32}x{30}|x{50}|x{50}x{60}}

\multirow{2}{*}{Dataset} & \multirow{2}{*}{\# views} & \multicolumn{3}{c}{Training time multiplier w.r.t. baseline} \\
& & \cellcolor{gray!25} NeRF \cite{mildenhall2020nerf} & +Ours & DietNeRF \cite{jain2021putting} \\
\shline
Blender & 8 & $1.0\times$ & $1.02\times$ & $2.8\times$\\
& & & & \\
Dataset & \# views &  \cellcolor{gray!25} mipNeRF \cite{barron2021mip} & +Ours & $^\dagger$RegNeRF \cite{niemeyer2022regnerf} \\
\shline
DTU & 3 & $1.0\times$ &  $1.04\times$ & $1.69\times$ \\
LLFF & 3 & $1.0\times$ & $1.04\times$ & $1.98\times$ \\
\end{tabular}
\vspace{-.5em}
\caption{\textbf{Training time comparison.} We run experiments under a fair setting and report the training time multipliers relative to the baselines. Our \OURS has negligible training overhead compared to baselines ({gray}), while DietNeRF and RegNeRF do not. $^\dagger$: w/o. appearance regularization. Note that using appearance regularization will further increase training budgets.}
\vspace{-1em}
\label{tab:overhead}
\end{table}

\subsection{Ablation Study}

In this section, we ablate our design choices on the DTU dataset and the LLFF dataset under the 3-view setting. We use a batch size of 1024 for faster training instead of 4096 for the main experiments in \Cref{tab:dtu,tab:llff}. 

\begin{figure}[t!]\centering
\includegraphics[width=\linewidth]{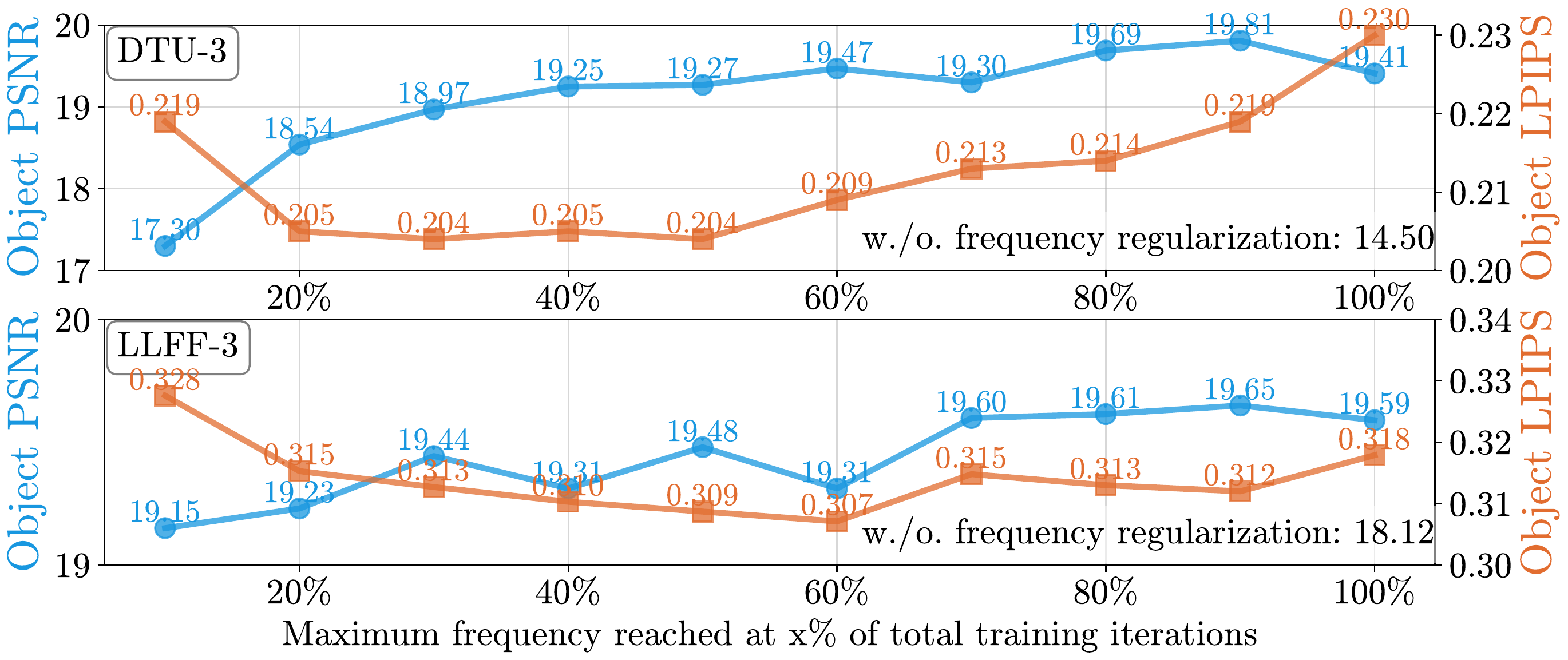}
\vspace{-1.5em}
\caption{\textbf{Effect of frequency regularization duration.} We set the end of frequency regularization as $T=\lfloor \mathrm{total\_iters} * x\%\rfloor$. \OURS achieves reasonably well performance across a wide range of curriculum choices. All entries use the occlusion regularization, including ``w/o. frequency regularization''.}
\vspace{-1em}
\label{fig:freq_curriculum}
\end{figure}

\paragraph{Frequency curriculum.}  We investigate the impact of frequency regularization duration $T$ in \Cref{fig:freq_curriculum}. Our \OURS benefits more from a longer curriculum in terms of PSNR score across two datasets, with the $90\%$-schedule being the best. We thus adopt it as our default schedule. However, we notice a trade-off between PSNR and LPIPS where a longer frequency regularization duration can result in higher PSNR but lower LPIPS scores. Fine-tuning the trained model can address this issue and yield better LPIPS scores. More details and discussions are provided in the Appendix.

\paragraph{Occlusion regularization.} \Cref{tab:ablation}-(a) studies the effect of occlusion regularization.
We observe consistent improvements in both datasets when occlusion regularization is included, confirming its efficacy. In contrast, the distortion loss $L_{distort}$ in \cite{barron2022mip} worsens the results. Additionally, we find the performance of DTU-3 drops significantly if a large $M$ is chosen since a large portion of real radiance fields falls in those ranges. The hyper-parameter $M$ can be set per dataset empirically according to the scene statistics. Further, in \Cref{tab:ablation}-(b), we show that the way our regularization penalizes points near the camera differs from simply adjusting the near bound. The latter changes the absolute location of the ray starting point, while the occlusion effect \textit{remains} in the starting area regardless of changes to the near bound.

\begin{table}[ht!]
\centering
\subfloat[\textbf{Ablation Study on $\mathcal{L}_{occ}$}.]{
\centering
\begin{minipage}{0.55\linewidth}{\begin{center}
\tablestyle{4pt}{1.05}
\resizebox{.9\linewidth}{!}{%
\begin{tabular}{y{80}x{30}x{30}}
\textbf{Settings (bs=1024)} & \textbf{DTU-3} & \textbf{LLFF-3} \\
\shline
\multicolumn{3}{l}{\textit{Ours v.s. occlusion regularization range $M$}} \\
w/t. $L_{distort}$ \cite{barron2022mip} & 15.14 & 19.08 \\ 
 w/o. $L_{\mathrm{occ}}$ & 17.40 & 19.16 \\
w/o. B\&W prior & 19.03 &  -- \\
B\&W prior only & 19.19 & -- \\
$M=5$ & \underline{19.78} & 19.24  \\
$M=10$  & \cellcolor{gray!25}\textbf{19.81} & 19.43 \\
$M=15$  & 18.57 & \underline{19.58} \\
$M=20$  & 13.76 & \cellcolor{gray!25}\textbf{19.70} \\
$M=25$  & 11.02 & 19.54 \\
\hline
\textbf{Ours} default (bs=1024) & 19.81 & 19.70 \\
\textbf{Ours} default (bs=4096) & \textbf{20.20} & \textbf{19.73} \\
\end{tabular}}
\end{center}}\end{minipage}

}
\subfloat[\textbf{$\mathcal{L}_{occ}$ v.s. near bounds}]{
\centering
\begin{minipage}{0.4\linewidth}{\begin{center}
\tablestyle{4pt}{1.05}
\resizebox{\linewidth}{!}{%
\begin{tabular}{x{20}x{35}x{35}}
\\
\textbf{Near} & w/o. $\mathcal{L}_{occ}$ & w/t. $\mathcal{L}_{occ}$ \\
\shline
\multicolumn{3}{l}{\textit{Ours v.s. tuning near bounds}} \\
0.0 & 17.40 & 19.09 \\
0.2 & 17.34 & 19.39 \\
0.4 & 17.43 & 19.61\\
\cellcolor{gray!25} 0.5 & 17.40 & \cellcolor{gray!25} \textbf{19.81} \\
0.6 & 17.35 & 19.11\\
0.7 & 16.73 & 19.11\\
0.8 & 15.08 & 16.77 \\
\\
\\
\end{tabular}}
\end{center}}\end{minipage}
}
\vspace{-.5em}
\caption{\textbf{Effect of occlusion regularization range.} (a) We report PSNR scores on the DTU-3 object and LLFF-3 datasets. Entries except the last row use a batch size of 1024. 
``B\&W'' means using the predicted black \& white color as additional prior (see ``Hyper-parameters'' in the ``Setup'' section). All entries use a $90\%$-schedule frequency regularization. (b) In the 3-view DTU ablation setting, we disable/enable $\mathcal{L}_{occ}$ and vary the near bound to study the impact of our occlusion regularization. Our results show consistent improvement while adjusting the near bound has little impact. Our default settings are marked in gray. }
\label{tab:ablation}
\vspace{-1em}
\end{table}

\paragraph{Limitations.} Our \OURS has two limitations. First, a longer frequency curriculum can make the scene smoother but may decrease LPIPS scores despite achieving competitive PSNR scores. Second, occlusion regularization can cause over-regularization and incomplete representations of near-camera objects in the DTU dataset. Per-scene tuning regularization range can alleviate this issue but we opt not to use it in this paper. Further discussion on these limitations can be found in the Appendix. Addressing these limitations can significantly improve \OURS and we leave them as future work. Still, we consider \OURS to be a simple yet intriguing \textit{baseline} approach for few-shot neural rendering that differs from the current trend of constructing more intricate pipelines.

\section{Conclusion}

We have presented FreeNeRF, a streamlined approach to few-shot neural rendering. Our study unfolds the deep relation between the input frequency and the failure of few-shot neural rendering. A simple frequency regularizer can drastically address this challenge. \OURS outperforms the existing state-of-the-art methods on multiple datasets with minimal overhead. Our results suggest several venues for future investigation. For example, it is intriguing to apply FreeNeRF to other problems suffering from high-frequency noise, such as NeRF in the wild \cite{martin2021nerf}, in the dark \cite{mildenhall2019local}, and even more challenging images in the wild, such as those from autonomous driving scenes. In addition, in the Appendix, we show that the frequency-regularized NeRF produces smoother normal estimation, which can facilitate applications that deal with glossy surfaces, as in RefNeRF~\cite{verbin2022ref}. We hope our work will inspire further research in few-shot neural rendering and the use of frequency regularization in neural rendering more generally.

{\small
\bibliographystyle{ieee_fullname}
\bibliography{egbib}
}

\clearpage
\appendix
\setcounter{table}{0}
\renewcommand{\thetable}{A.\arabic{table}}
\setcounter{figure}{0}
\renewcommand{\thefigure}{A.\arabic{figure}}

\twocolumn[{%
\centering

{\setstretch{1.25} \Large \textbf{Supplement to FreeNeRF: Improving Few-shot Neural Rendering with Free Frequency Regularization} \par}
\vspace{1em}
Project page: \href{https://jiawei-yang.github.io/FreeNeRF/}{FreeNeRF}
\vspace{2em}

\includegraphics[width=\textwidth,height=\textheight,keepaspectratio]{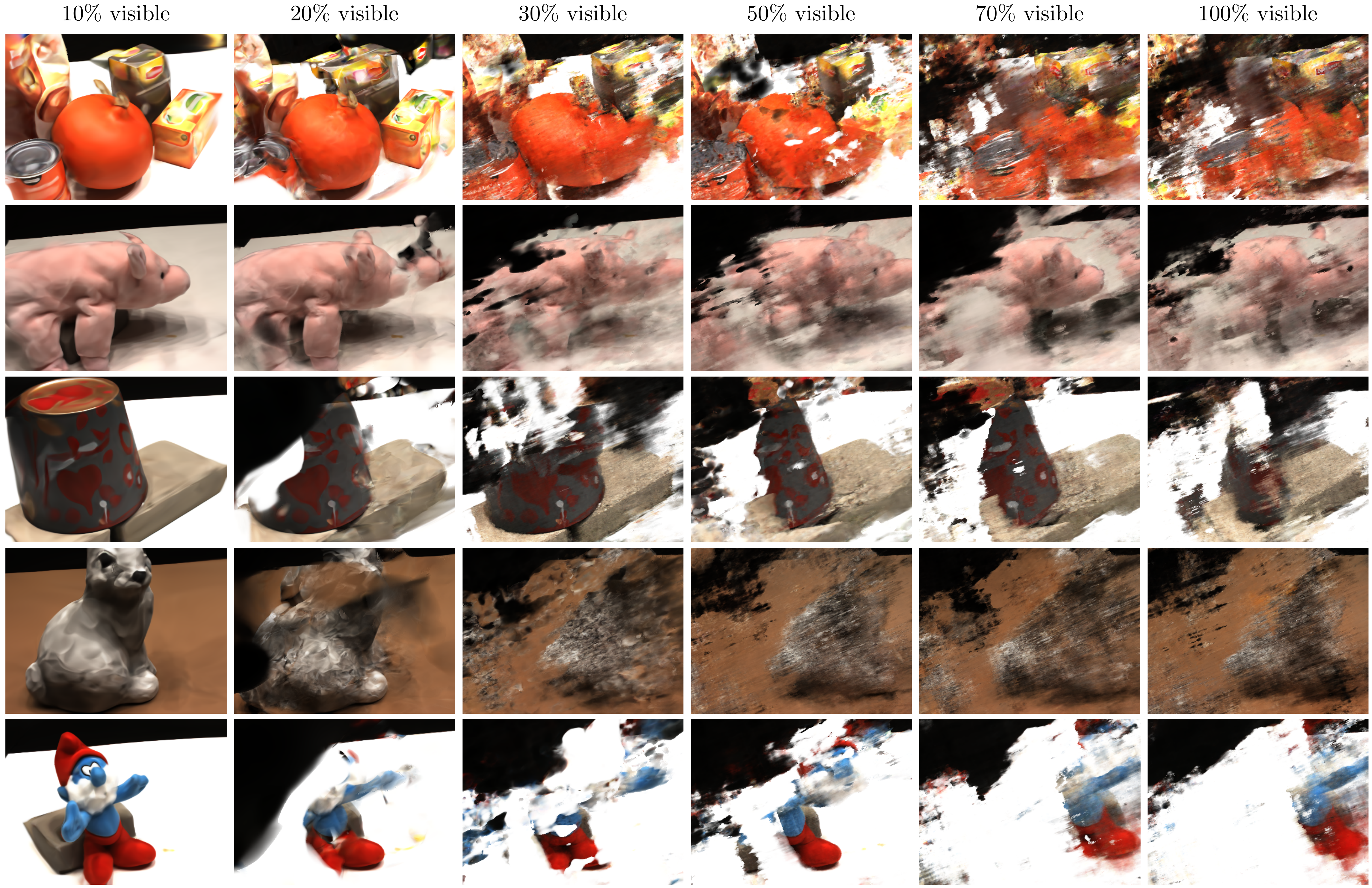}
\captionof{figure}{\textbf{High-frequency inputs cause catastrophic failure in few-shot neural rendering.} We train mipNeRF \cite{barron2021mip} with masked (integrated) positional encoding by setting \texttt{pos\_enc[int(L*x\%]):]=0}, where $L$ denotes the length of frequency bands (\cref{eq:pos_enc}) and $x$ is the masking ratio. Using low-frequency components as inputs enables mipNeRF to learn meaningful scene representations despite their over-smoothness. Please refer to \Cref{fig:masking} (in the main text) for numerical comparisons. We also provide animated visualizations on our project page.}
\label{fig:max_freq}
\vspace{1em}
}]

In this supplement, we include additional quantitative and qualitative results to discuss more motivation and limitations of \OURS in \Cref{sec:add_results}. We also add details of experimental settings and implementations in \Cref{sec:exp_details}.

\section{Additional Results}
\label{sec:add_results}

\paragraph{High-frequency inputs cause catastrophic failure.} \Cref{fig:max_freq} shows more examples to demonstrate the failure mode revealed in \Cref{fig:masking} that the high-frequency inputs lead to the catastrophic failure of few-shot neural rendering. When taking in 10\% of the total embedding bits, mipNeRF can successfully reconstruct scenes despite their over-smoothness. However, with higher-frequency inputs, the scene reconstructions become more unrecognizable and collapse. This experimental finding lies at the heart of \OURS: by restricting the inputs to the low-frequency components at the start of training, NeRF can start from significantly stabilized scene representations at the early stage of training. Upon these stable scene representations,  NeRF continues refining the details when high-frequency signals become visible.

\subsection{Limitations}
\label{sec:limitation}
In this subsection, we elaborate on the limitations and showcase the failure cases of \OURS.

\begin{figure}[t]\centering
\includegraphics[width=\linewidth]{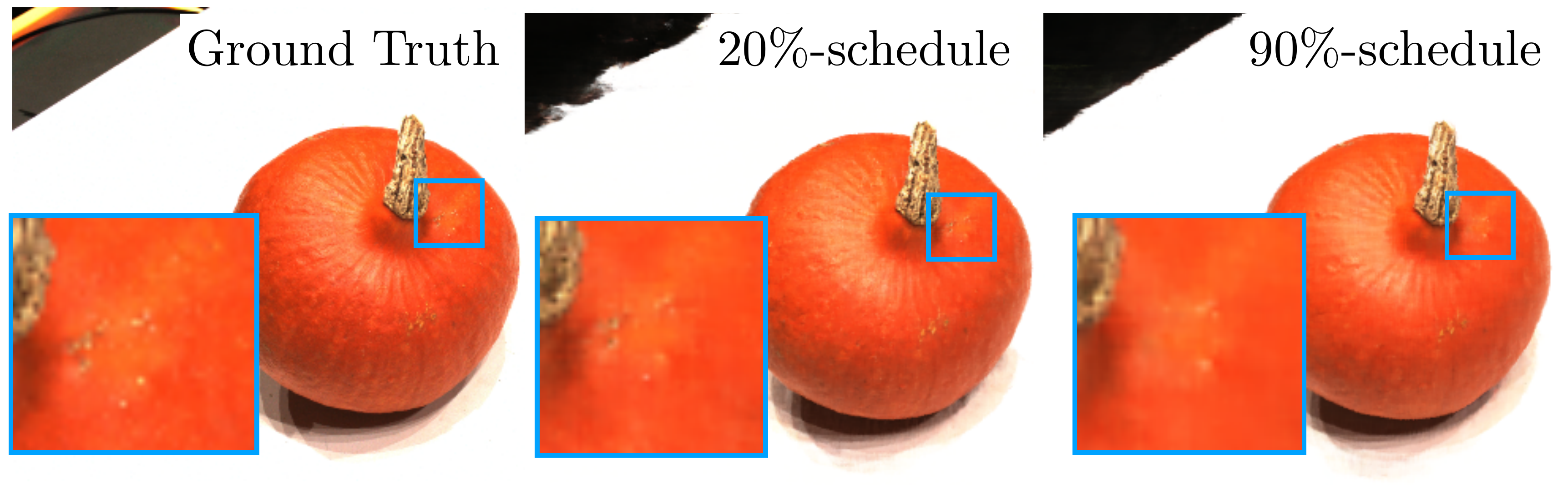}
\vspace{-2em}
\caption{\textbf{High-frequency details comparison.} We show the view synthesis results under the 9 input-view setting on the DTU dataset. With enough view information, a shorter frequency regularization enables NeRF models to render more high-frequency details.}
\label{fig:high_freq}
\end{figure}

\paragraph{Trade-off between PSNR and LPIPS.} \Cref{fig:freq_curriculum} studies the effect of the duration of frequency regularization on PSNR and LPIPS. From the figure, we observe a trade-off between PSNR and LPIPS that a long-frequency curriculum usually results in a high PSNR score but a low LPIPS score. For example, under the 9 input-view setting, we obtain an object PSNR of 25.59 and an object LPIPS of 0.117 with a 90\%-schedule and those of 25.38 and 0.096 with a 50\%-schedule. Visually, when the number of input views is relatively sufficient (but still under few-shot settings), results under a shorter schedule usually present more high-frequency details (see the zoom-in patch in \cref{fig:high_freq}). 
We thus use 70\%-schedule and 50\%-schedule for experiments under 6 and 9 input-view settings, respectively. We also found out that training FreeNeRF longer can obtain better LPIPS performance, \emph{e.g.}, 0.182 to 0.167 and 0.308 to 0.290 for DTU-3 and LLFF-3 settings, respectively.

\begin{figure}[h]\centering
\includegraphics[width=\linewidth]{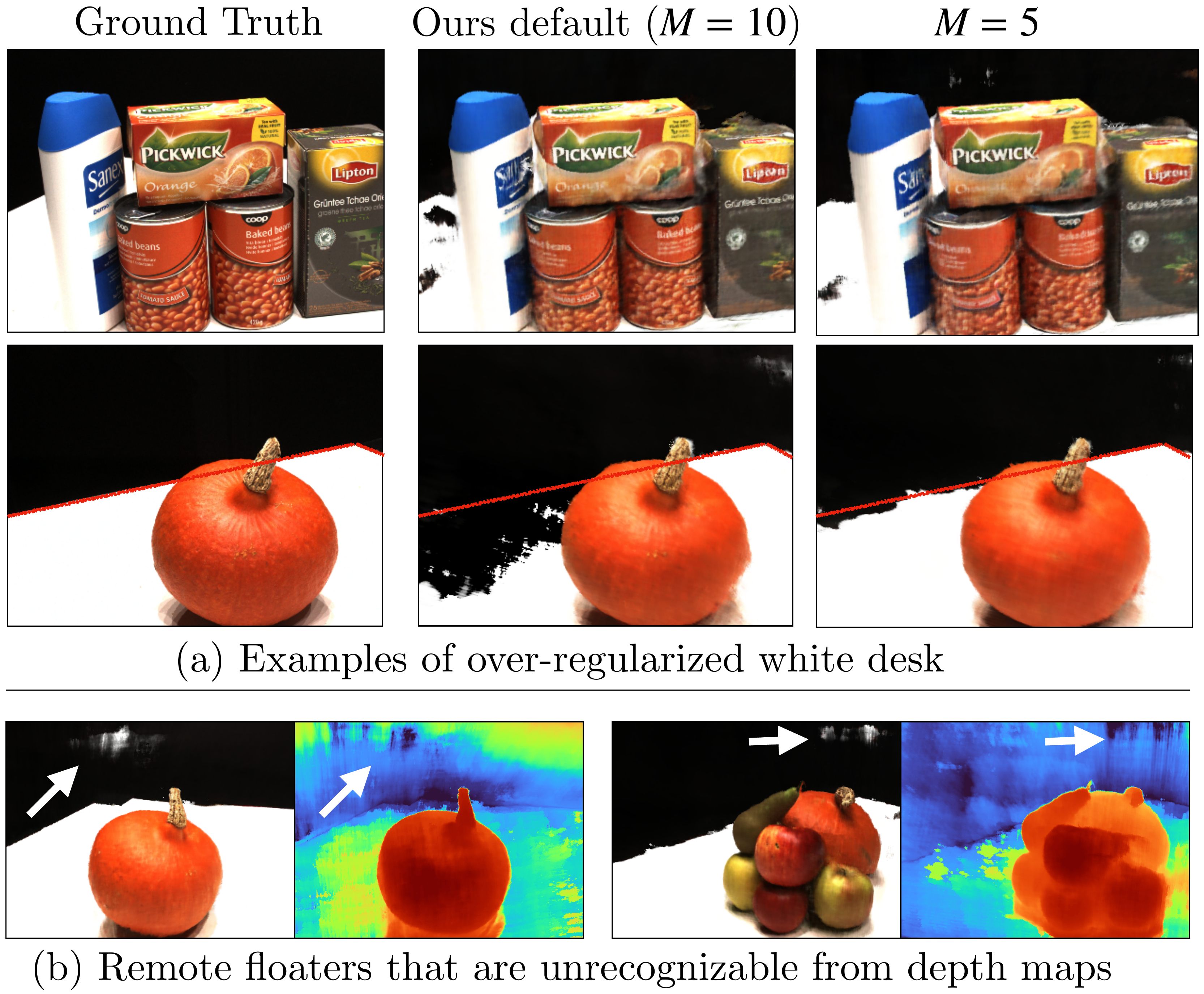}
\vspace{-1em}
\caption{\textbf{Limitations of occlusion regularization.} (a) Aggressive occlusion regularization results in incomplete white desks that are visually annoying. Reducing the regularization range (from $M=10$ to $M=5$) can alleviate the issue to some extent. (b) Occlusion regularization does not solve remote floaters that are far from cameras.}
\label{fig:over_reg}
\vspace{-1em}
\end{figure}

\paragraph{Limitations of $L_{\mathrm{occ}}$.} \textbf{Over-regularization:} our occlusion regularization can lead to an incomplete white desk on the DTU dataset due to over-regularization in some scenes, as shown in \Cref{fig:over_reg}-(a). Reducing the regularization range of $L_{\mathrm{occ}}$ can ease this issue. A set of per-scene tuned hyper-parameters can potentially provide better results. \textbf{Remote floaters:} \Cref{fig:over_reg}-(b) shows some small cloudy floaters far from the camera. Our occlusion regularization that penalizes near-camera dense fields does not solve this problem. However, we do not observe these remote floaters in NeRF trained with only low-frequency inputs (10\% visible). That said, though significantly regularized and stabilized, FreeNeRF still overfits to spurious occupancy to a certain degree. Better performance is excepted if  FreeNeRF further exploits the low-frequency components to avoid such overfitting, leaving room for future work and improvements.

\subsection{Depth Evaluation}
Here we include results to compare the capability of different methods in depth estimation. As the datasets do not have actual ground truth depth, we utilized depth maps generated by mipNeRFs that were trained on all views as a substitute. FreeNeRF significantly improves its baseline, mipNeRF. RegNeRF, with its patch-based geometry regularization, achieves better performance on the object-centric DTU dataset, while FreeNeRF performs better on the scene-scale LLFF dataset without explicit geometry regularization. This experiment demonstrates the different features of FreeNeRF and RegNeRF, as well as the differences between DTU and LLFF datasets.

\begin{center}
\tiny
\tablestyle{1pt}{1.1}
\begin{tabular}{y{95} | x{21} x{21} x{21} |x{21} x{21} x{21}}

Error=$\Vert D_{pseu} - D_{pred} \Vert$ & \multicolumn{3}{c}{DTU obj {depth error$\downarrow$}} & \multicolumn{3}{|c}{LLFF {depth error$\downarrow$}} \\
\hline
\# views & 3 & 6 & 9 & 3 & 6 & 9 \\
\shline
mipNeRF (baseline) & 131.97 & 59.21 & 18.73 & 149.18 & 36.92 & 19.16  \\
RegNeRF (explicit geo. reg.)  & 14.58 & 10.40 & 6.23 & 44.52 & 25.09 & 18.26 \\ 
FreeNeRF & 14.89 & 12.98 & 9.48 & 39.92 & 23.61 & 16.91 \\
\hline
\end{tabular}
\end{center}

\subsection{Additional Qualitative Results}

\Cref{tab:dtu_more} provides more numeric results in addition to \Cref{tab:dtu} on the DTU dataset. \OURS achieves the best results under the ``Average'' metrics in most settings. However, we observe less improvement in terms of LPIPS. As we analyze in \Cref{sec:limitation}, the slight blurriness introduced by \OURS will result in a low LPIPS score. This is a limitation that could be addressed in the future.

\begin{table*}
\centering
\small
\tablestyle{2pt}{1.1}
\begin{tabular}{l | c | x{24}x{24}x{24} | x{24}x{24}x{24} | x{24}x{24}x{24} | x{24}x{24}x{24} }
& \multirow{2}{*}{Setting} & \multicolumn{3}{c|}{Object LPIPS $\downarrow$} & \multicolumn{3}{c|}{Object Average $\downarrow$} & \multicolumn{3}{c|}{Full-image LPIPS $\downarrow$} & \multicolumn{3}{c}{Full-image Average $\downarrow$}  \\
  &  & 3-view & 6-view & 9-view  & 3-view & 6-view & 9-view  & 3-view & 6-view & 9-view  & 3-view & 6-view & 9-view \\ \shline
SRF~\cite{chibane2021stereo} & \multirow{3}{*}{Trained on DTU} & 0.304 & 0.250 & 0.232 & 0.171 & 0.132 & 0.120 & 0.482 & 0.401 & 0.359 & 0.207 & 0.162 & 0.145 \\
PixelNeRF~\cite{yu2021pixelnerf} &  & 0.270 & 0.232 & 0.220 & 0.147 & 0.115 & 0.100 & 0.401 & 0.340 & 0.323 & 0.154 & \cellcolor{yellow!25} 0.119 & 0.105 \\
MVSNeRF~\cite{chen2021mvsnerf} &  &0.197 & 0.156 & 0.135 &  \cellcolor{yellow!25}0.113 &  0.088 & 0.068 &  0.385 &  0.321 &  0.280 &  \cellcolor{orange!25} 0.184 &  0.146 & 0.114 \\
\hline 
SRF ft~\cite{chibane2021stereo} & \multirow{3}{*}{\shortstack{Trained on DTU\\and\\Optimized per Scene}} & 0.281 & 0.225 & 0.205 & 0.162 & 0.114 & 0.093 & 0.431 & 0.353 & 0.325 & 0.196 & 0.143 & 0.125  \\
PixelNeRF ft~\cite{yu2021pixelnerf} &  &  0.269 & 0.223 & 0.203 & 0.125 & 0.104 & 0.090 & 0.456 & 0.351 & 0.338 & \cellcolor{yellow!25} 0.185 & 0.121 & 0.117 \\
MVSNeRF ft~\cite{chen2021mvsnerf} &  & 0.197 &  0.155 &  0.135 &  \cellcolor{yellow!25}0.113 &  0.089 & 0.069 &  0.384 & 0.319 & 0.278 & \cellcolor{yellow!25} 0.185 & 0.146 & 0.113 \\
\hline 
mip-NeRF~\cite{barron2021mip} & \multirow{3}{*}{Optimized per Scene} & 0.353 & 0.198 &  \cellcolor{yellow!25}0.092 & 0.323 & 0.148 &  \cellcolor{yellow!25}0.056 & 0.655 & 0.394 & 0.209 & 0.485 & 0.231 & 0.098\\
DietNeRF~\cite{jain2021putting} &  & 0.314 & 0.201 & 0.173 & 0.243 & 0.101 & 0.068 & 0.574 & 0.336 & 0.277 & 0.383 & 0.149 & 0.098 \\
RegNeRF~\cite{niemeyer2022regnerf} &  & \cellcolor{orange!25}0.190 &  \cellcolor{red!25}0.117 & \cellcolor{orange!25} 0.089 &  \cellcolor{orange!25}0.112 &  \cellcolor{yellow!25}0.071 &  \cellcolor{orange!25}0.047 &  \cellcolor{orange!25} 0.341 & \cellcolor{red!25} 0.233 & \cellcolor{orange!25} 0.184 & 0.189 & \cellcolor{orange!25} 0.118 & \cellcolor{yellow!25} 0.079 \\ \hline

mip-NeRF concat. (repro.) & \multirow{3}{*}{Optimized per Scene} & \cellcolor{gray!25} 0.348 & \cellcolor{gray!25} 0.197 & \cellcolor{gray!25} 0.100 & \cellcolor{gray!25} 0.311 &  \cellcolor{gray!25} 0.144 & \cellcolor{gray!25} 0.057 & \cellcolor{gray!25} 0.643 & \cellcolor{gray!25} 0.403 & \cellcolor{gray!25} 0.218 & \cellcolor{gray!25} 0.472 & \cellcolor{gray!25} 0.240 & \cellcolor{gray!25} 0.099 \\
$^\dagger$RegNeRF concat. (repro.) &   & \cellcolor{yellow!25}0.196 &\cellcolor{orange!25} 0.118 & \cellcolor{red!25} 0.088 & 0.117  & \cellcolor{orange!25} 0.070 & \cellcolor{red!25} 0.046 & \cellcolor{yellow!25} 0.350 & \cellcolor{orange!25} 0.236 & \cellcolor{red!25} 0.183 &  0.197 & \cellcolor{orange!25} 0.118 & \cellcolor{orange!25} 0.078  \\
\textbf{Our \OURS} &  & \cellcolor{red!25} 0.182 & \cellcolor{yellow!25} 0.137 & 0.096 & \cellcolor{red!25} 0.098 & \cellcolor{red!25} 0.068 & \cellcolor{orange!25} 0.046 & \cellcolor{red!25} 0.318 &  \cellcolor{yellow!25} 0.240 & \cellcolor{yellow!25} 0.187 & \cellcolor{red!25} 0.146 &\cellcolor{red!25} 0.094 & \cellcolor{red!25} 0.068 \\
\end{tabular}
\caption{
    \textbf{Quantitative comparison on DTU.} We provide additional quantitative results to \Cref{tab:dtu}. Results in the bottom row are reproduced by us, and others come from \cite{niemeyer2022regnerf}. ``concat.'': inputs concatenation (\cref{eq:concat}). $^\dagger$ReNeRF: w/o. appearance regularization. The best, second-best, and third-best entries are marked in red, orange, and yellow, respectively.
    }
\label{tab:dtu_more}
\end{table*}

\subsection{Additional Visualizations}
\begin{figure*}[t]\centering
\includegraphics[width=0.9\textwidth]{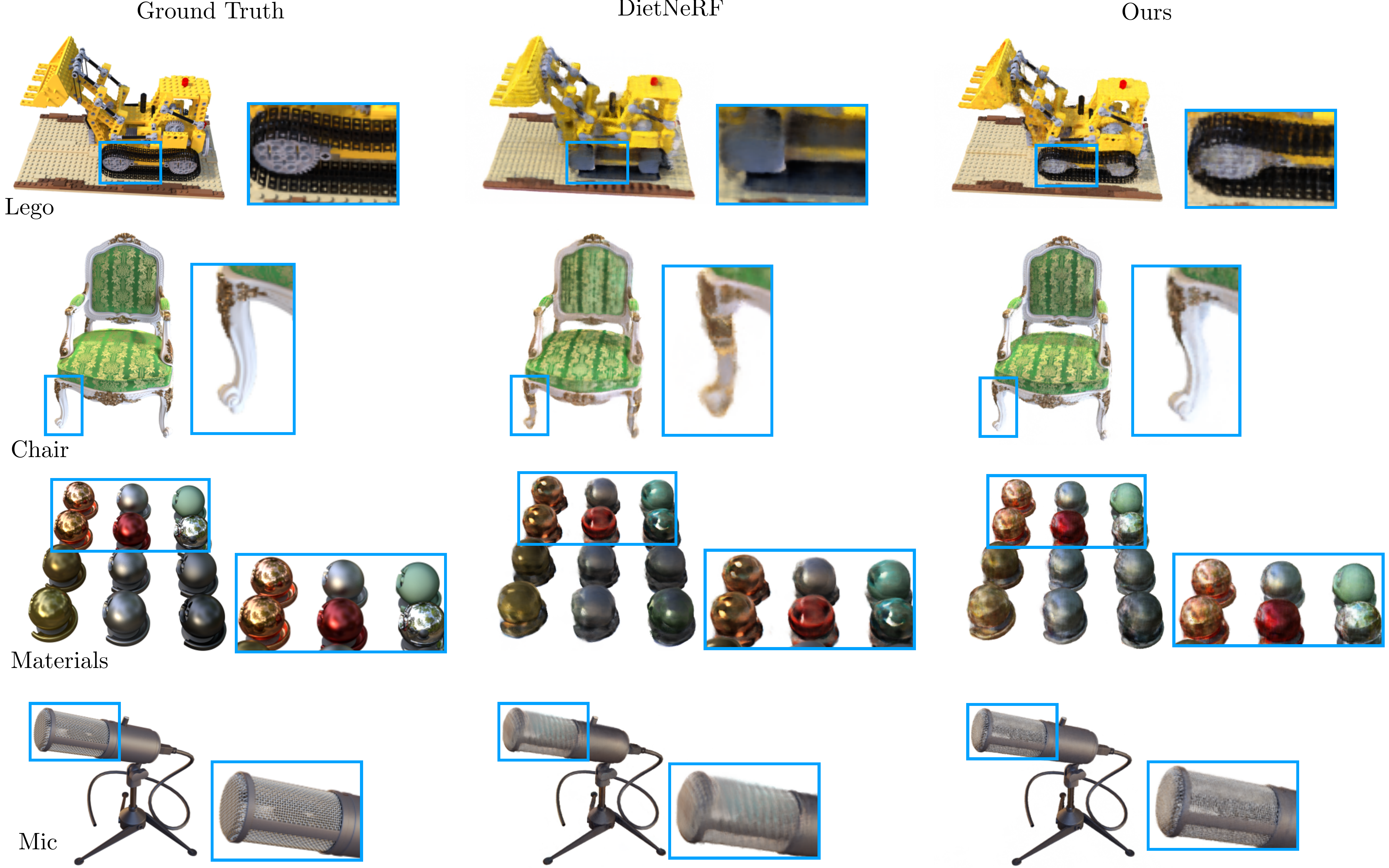}
\caption{\textbf{Qualitative comparison on the Blender dataset.} DietNeRF generates patches that can be reasonable and plausible to some extent but do not closely match the ground truth. This is a limitation of using a pre-trained model for semantic regularization. In contrast, our FreeNeRF reconstructs scenes that are more in line with the ground truth.}
\label{fig:dietnerf_more}
\end{figure*}

\paragraph{Blender.} In \Cref{fig:dietnerf_more}, we show more qualitative comparisons between DietNeRF \cite{jain2021putting} and our FreeNeRF on the Blender dataset. From the zoom-in patches of DietNeRF's results, we see the generated patches are blurry and do not reflect the same distribution of style as that of ground truth. This is due to implicit semantics distillation behavior done by DietNeRF. In contrast, our FreeNeRF reconstructs scenes closer to the ground truth.

\begin{figure*}[t]\centering
\includegraphics[width=\textwidth,height=\textheight,keepaspectratio]{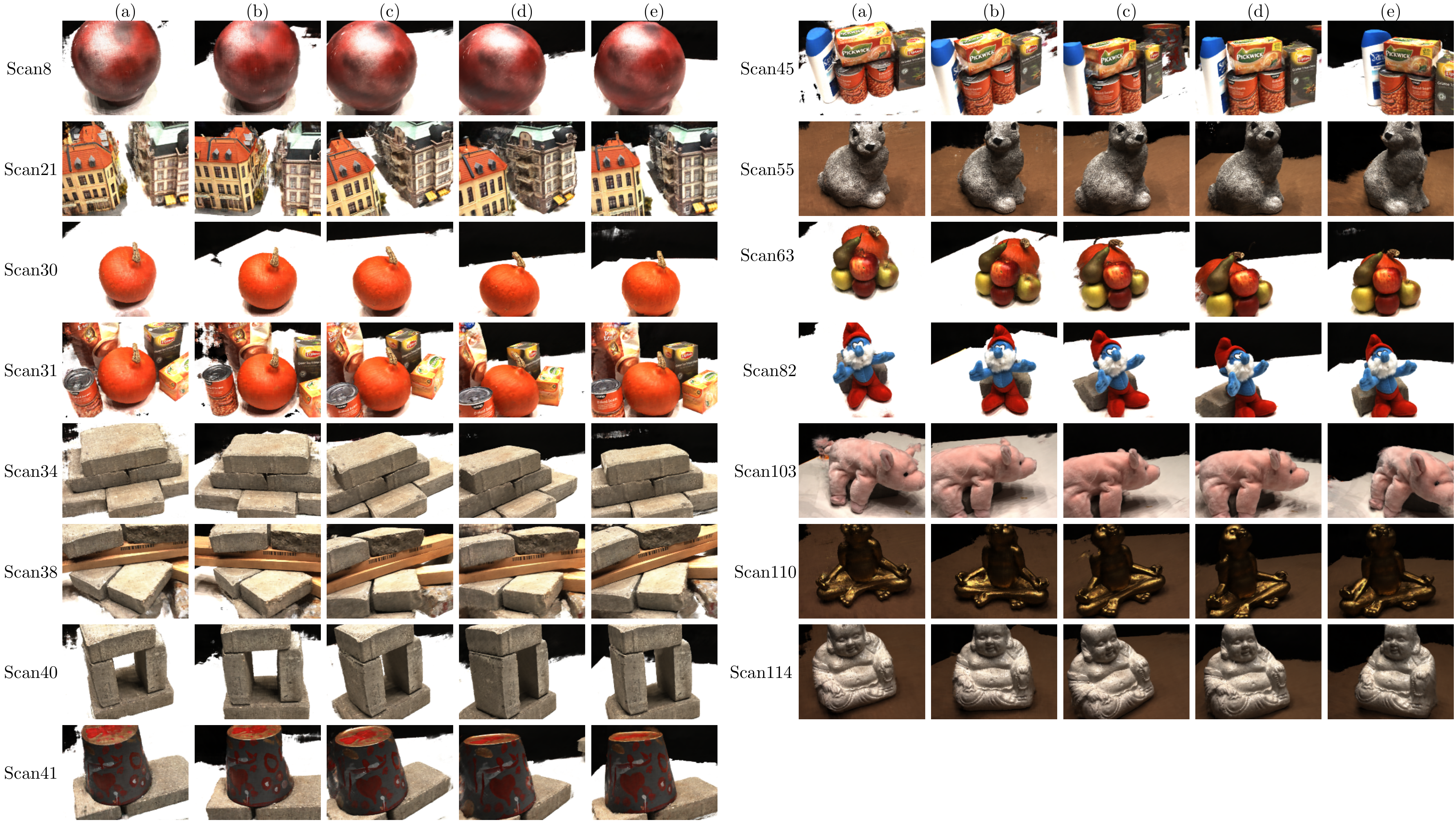}
\vspace{-2em}
\caption{\textbf{Example \OURS's novel view synthesis results with 3 input views on the DTU dataset.}}
\vspace{-1em}
\label{fig:dtu3_more}
\end{figure*}

\begin{figure*}[t]\centering
\includegraphics[width=\textwidth,height=\textheight,keepaspectratio]{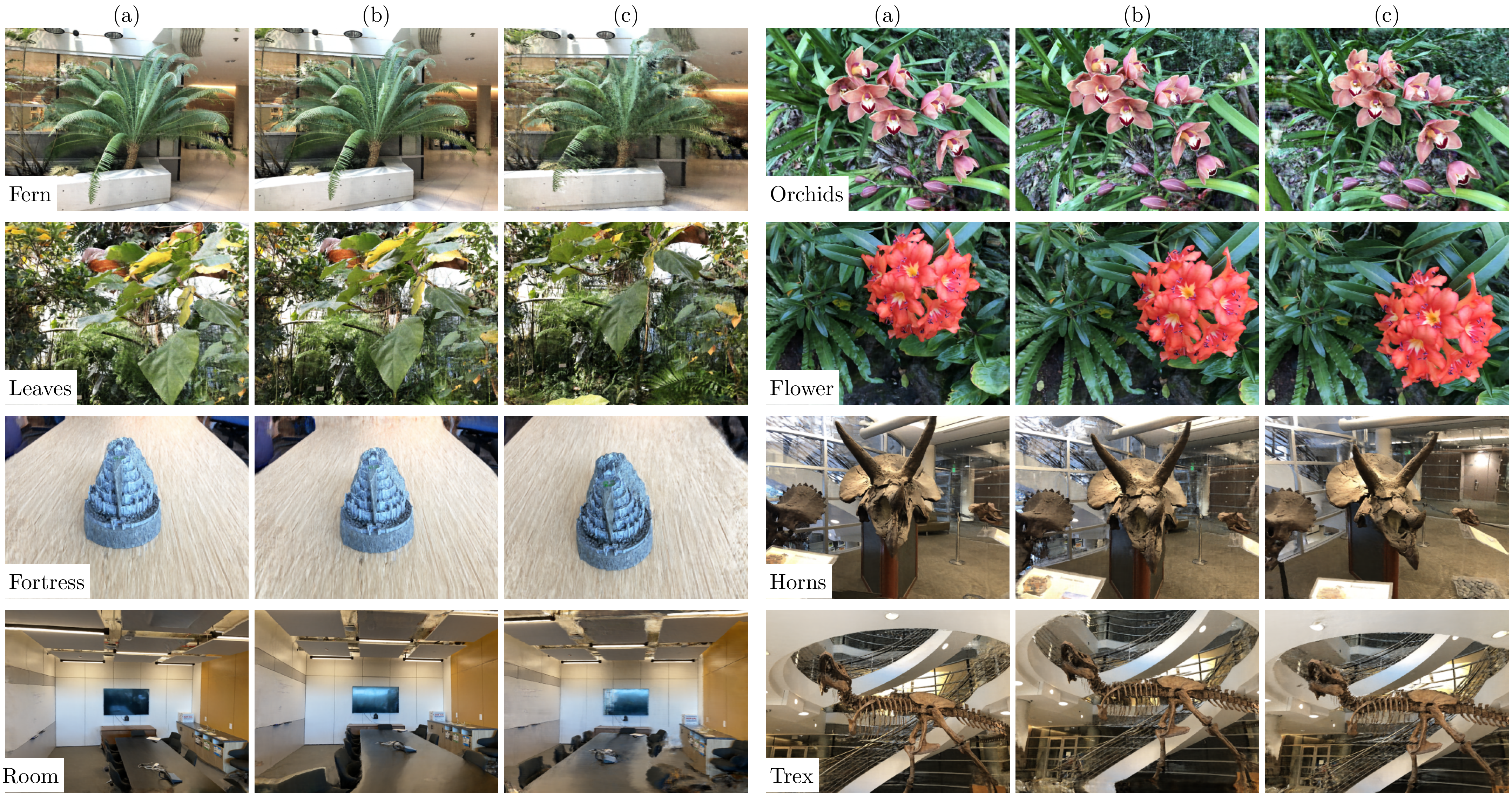}
\vspace{-2em}
\caption{\textbf{Example \OURS's novel view synthesis results with 3 input views on the LLFF dataset.}}
\vspace{-1em}
\label{fig:llff3_more}
\end{figure*}

\paragraph{DTU and LLFF.} We provide more rendering results by FreeNeRF in \Cref{fig:dtu3_more,fig:llff3_more} under the 3 input-view setting on the DTU dataset and the LLFF dataset, respectively.

\subsection{FreeNeRF for Normal Estimation}
\begin{figure*}[t]\centering
\includegraphics[width=\textwidth,height=\textheight,keepaspectratio]{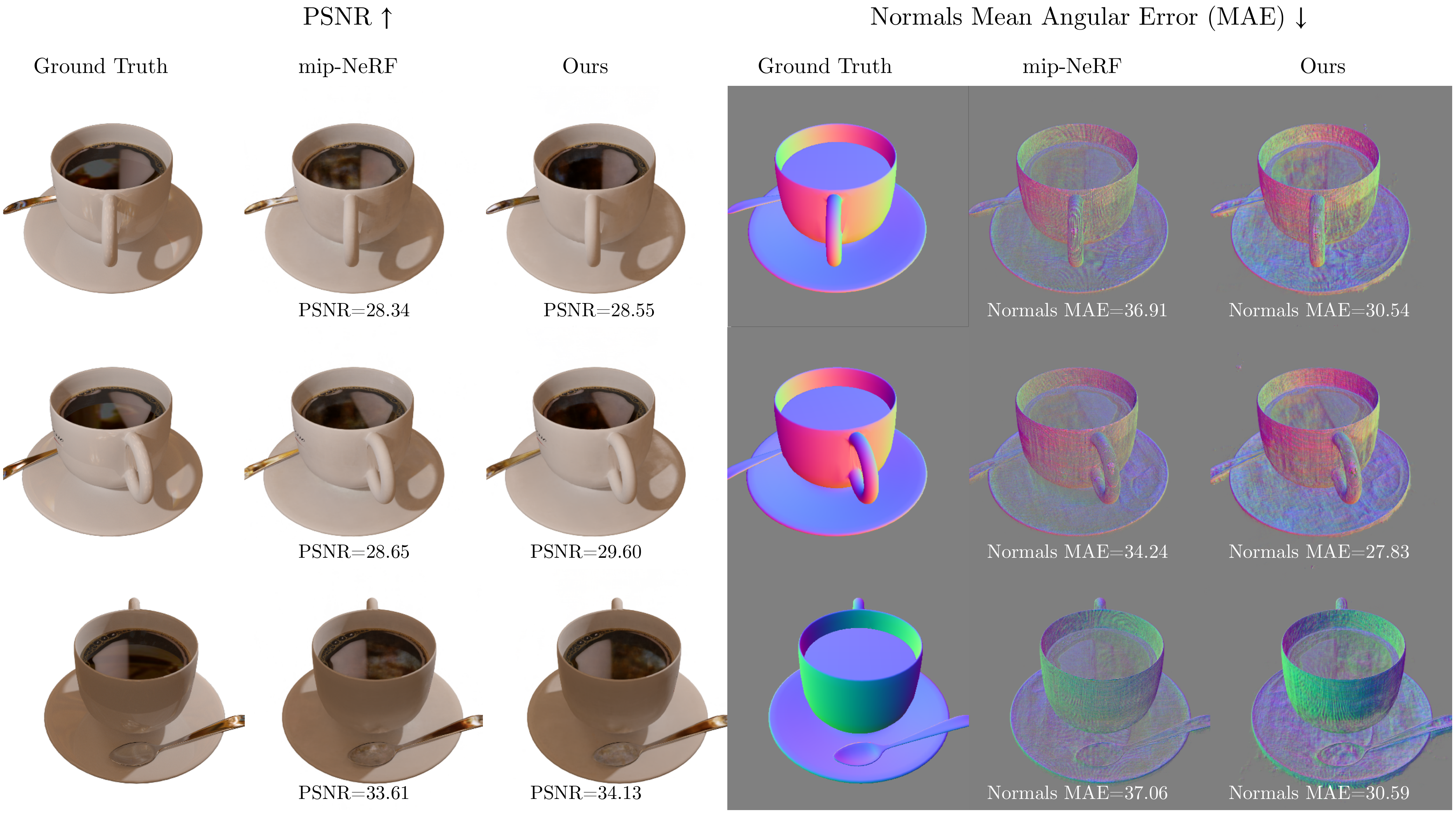}
\vspace{-2em}
\caption{\textbf{Comparison on normal vectors estimation.} Beyond the few-shot neural rendering problem, we train a mipNeRF and a FreeNeRF on the ``coffee'' scene in the Shiny Blender dataset \cite{verbin2022ref} to demonstrate FreeNeRF's potential in estimating more accurate normal vectors. The PSNR scores for this scene are 30.839 and 31.364 for mipNeRF and FreeNeRF, respectively. The mean angular errors (the lower, the better) are 36.549 and 31.492 for mipNeRF and FreeNeRF, respectively. Note that we use a much smaller batch size (4096) than that in the original setting (16394), so the numerical results here are not comparable to those in RefNeRF \cite{verbin2022ref}.}
\label{fig:normals}
\end{figure*}

We briefly demonstrate a potential FreeNeRF's application beyond few-shot neural rendering. Specifically, we follow the similar settings in RefNeRF\cite{verbin2022ref} to train a mipNeRF and a FreeNeRF on the ``coffee'' scene in the Shiny Blender dataset \cite{verbin2022ref}. This dataset aims to benchmark NeRF's performance on glossy surfaces, where the key challenge is to estimate accurate normal vectors. \Cref{fig:normals} shows the comparison between mipNeRF and FreeNeRF. Compared to mipNeRF, FreeNeRF produces more accurate normal estimation and achieves much lower mean angular error (MAE) at no sacrifice of PSNR score. We conjecture that overfitting to high-frequency signals at the start of training is a very common issue in NeRF's training. However, such partial failure is veiled by good appearance results. We believe these partially degenerated results can be improved with frequency regularization, which makes NeRF's initial training more stable.

\vspace{-1em}
\section{Experiment Details}
\label{sec:exp_details}

We strictly follow the experimental settings in DietNeRF \cite{jain2021putting} and RegNeRF \cite{niemeyer2022regnerf} to conduct our experiments. We provide some details in the following for completeness.

\subsection{Dataset and metrics.}

\paragraph{Blender Dataset.} The Blender dataset \cite{mildenhall2020nerf} has 8 synthetic scenes in total. We follow the data split used in DietNeRF \cite{jain2021putting} to simulate a few-shot neural rendering scenario. For each scene, the training images with IDs (counting from ``0'') 26, 86, 2, 55, 75, 93, 16, 73, and 8 are used as the input views, and 25 images are sampled evenly from the testing images for evaluation. We follow \cite{jain2021putting} to use a $2\times$ downsampled resolution, resulting in $400\times 400$ pixels for each image.

\paragraph{DTU Dataset.} The DTU dataset \cite{jensen2014large} is a large-scale multi-view dataset that consists of 124 different scenes. PixelNeRF \cite{yu2021pixelnerf} uses a split of 88 training scenes and 15 test scenes to study the ``{pre-training \& per-scene fine-tuning}'' setting in a few-shot neural rendering scenario. Different from theirs, our method does not require pre-training. We follow \cite{niemeyer2022regnerf} to optimize NeRF models directly on the 15 test scenes. The test scan IDs are: 8, 21, 30, 31, 34, 38, 40, 41, 45, 55, 63, 82, 103, 110, and 114. In each scan, the images with the following IDs (counting from ``0'') are used as the input views: 25, 22, 28, 40, 44, 48, 0, 8, 13. The first 3 and 6 image IDs correspond to the input views in 3- and 6-view settings, respectively. The images with IDs in [1, 2, 9, 10, 11, 12, 14, 15, 23, 24, 26, 27, 29, 30, 31, 32, 33, 34, 35, 41, 42, 43, 45, 46, 47] serve as the novel views for evaluation. The remaining images are excluded due to wrong exposure. We follow \cite{niemeyer2022regnerf,yu2021pixelnerf} to use a 4$\times$ downsampled resolution, resulting in $300\times 400$ pixels for each image.

 \paragraph{LLFF Dataset.} The LLFF dataset \cite{mildenhall2019local} is a forward-facing dataset that contains 8 scenes in total. Adhere to \cite{niemeyer2022regnerf,mildenhall2020nerf}, we use every 8-th image as the novel views for evaluation, and evenly sample the input views across the remaining views. Images are downsampled $8\times$, resulting in $378\times 504$ pixels for each image.

\paragraph{Metrics.} To compute PSNR scores, we use the formula $-10\cdot\log_{10}(\mathrm{MSE})$ (assuming the maximum pixel value is 1). Additionally, we utilize the scikit-image's API\footnote{\scriptsize\url{https://scikit-image.org/docs/stable/auto_examples/transform/plot_ssim.html}} to compute the structural similarity index measure (SSIM) score and the interface provided by an open source repository\footnote{\scriptsize\url{https://github.com/richzhang/PerceptualSimilarity}} (using a learned VGG model) to compute the learned perceptual image patch similarity (LPIPS) score.

\subsection{Implementations.}

\paragraph{DietNeRF's codebase.} In this codebase\footnote{\scriptsize\url{https://github.com/ajayjain/DietNeRF}}, a plain NeRF \cite{mildenhall2020nerf} that consists of two MLPs (one coarse MLP and one fine MLP) is used as the baseline. All NeRF models are trained with the Adam optimizer for 200k iterations. The learning rate starts at $5\times10^{-4}$ and decays exponentially with a rate of 0.1. We refer readers to the codebase for more details. 
In this codebase, the maximum input frequency $L$ (\cref{eq:pos_enc}) used in the position encoding for coordinates is $9$.
The original coordinates are concatenated with positional encodings by default.

\paragraph{RegNeRF's codebase.} In this codebase\footnote{\scriptsize\url{https://github.com/google-research/google-research/tree/master/regnerf}}, a plain mipNeRF \cite{barron2021mip} is used as the baseline. The maximum input frequency of coordinates is $16$, which is larger than that of the original NeRF \cite{mildenhall2020nerf}. We further concatenate the original coordinates into the positional encodings. All NeRF models are trained with the Adam optimizer with an exponential learning rate decaying from $2\cdot 10^{-3}$ to $2\cdot 10^{-5}$ and 512 warm-up steps with a multiplier of 0.01 \cite{barron2021mip}. Following \cite{niemeyer2022regnerf}, we clip gradients by value at 0.1 and then by norm at 0.1 for all experiments. All NeRF models in the main experiments are optimized for 500 epochs with a batch size of 4096. This setting results in around 44k, 88k and 132k training iterations on the DTU dataset for 3/6/9 input views, respectively, and 70k, 140k and 210k training iterations for those on the LLFF dataset, respectively. Note that in the ablation study we use a batch size of 1024 instead of 4096 for faster training.

\paragraph{Occlusion regularization.} We use a weight of $0.01$ for $L_{\mathrm{occ}}$ in all experiments. For simplicity, we compute this loss on the secondary stage's outputs, \ie those from the fine MLP in NeRF \cite{mildenhall2020nerf} and the second query in mipNeRF \cite{barron2021mip}.

\end{document}